\documentclass[12pt]{article}
\usepackage{amsmath}
\usepackage{times}
\usepackage{graphicx}
\usepackage{color}
\usepackage{multirow}
\usepackage{placeins}
\usepackage{float}
\usepackage{mathptmx} 
\usepackage[T1]{fontenc}
\usepackage{makecell}
\usepackage{booktabs}
\usepackage{tabularx}
\usepackage{amsmath, amssymb}
\usepackage{array}

\usepackage{tabularx,booktabs}
\usepackage{ragged2e}

\usepackage{natbib}
\usepackage{etoolbox}
\usepackage{titlesec}
\titleformat{\section}{\normalfont\normalsize\bfseries}{\thesection}{1em}{}
\titleformat{\subsection}{\normalfont\normalsize\bfseries}{\thesubsection}{1em}{}
\titleformat{\subsubsection}{\normalfont\normalsize\bfseries}{\thesubsubsection}{1em}{}

\usepackage{microtype}
\usepackage{graphicx}
\usepackage{bbm}
\usepackage{latexsym}
\usepackage{hyperref}
\hypersetup{hidelinks}

\renewcommand{\thesubsubsection}{\arabic{section}.\arabic{subsubsection}}

\makeatletter  
\patchcmd{\@footnotetext}{\reset@font\footnotesize}{\reset@font\normalsize}{}{}
\patchcmd{\@mpfootnotetext}{\reset@font\footnotesize}{\reset@font\normalsize}{}{}
\long\def\@makecaption#1#2{%
  \vskip\abovecaptionskip
  {\normalsize\noindent #1: #2\par}%
  \vskip\belowcaptionskip}
\makeatother   

\begin{document}
\pagestyle{plain}

\ \vspace{20mm}\\

\begin{center}{\bfseries Behavioral Latency as Weak Event-Time Supervision for EEG Reaction-Time Decoding}
\end{center}

\ \\
{\bf Anuar Aimoldin$^{1}$\textsuperscript{*}}\\
anuar.aimoldin@mbzuai.ac.ae

\ \\ {\bf Ayana Mussabayeva$^{1}$\textsuperscript{*}}\\
ayana.mussabayeva@mbzuai.ac.ae

\ \\ {\bf Yedige Mussabayev$^{\displaystyle 2}$}\\
ymussabayev@hof-university.de

\ \\ {\bf Xue Liu$^{\displaystyle 1, 3}$}\\
steve.liu@mbzuai.ac.ae

\ \\ {\bf Kun Zhang$^{\displaystyle 1, 4}$}\\
kun.zhang@mbzuai.ac.ae

\ \\
{$^{\displaystyle 1}$Mohamed bin Zayed University of Artificial Intelligence (MBZUAI), Abu Dhabi, UAE}
\ \\
{$^{\displaystyle 2}$ Hof University, Hof, Germany}
\ \\
{$^{\displaystyle 3}$ McGill University, Montreal, QC, Canada}
\ \\
{$^{\displaystyle 4}$ Carnegie Mellon University, Pittsburgh, PA, USA}
\ \\
{\normalsize * Equal contribution}\\

\noindent\emph{This is the authors' final version of the manuscript accepted for publication in \textit{Neural Computation}.}

\ \\[-2mm]

\ \vspace{-0mm}\\
\begin{center} {\bf Abstract} \end{center}

Single-trial EEG analyses are often organized around events and latencies, such as stimulus onset, response execution, ERP component timing, and movement onset. Yet EEG-based reaction-time prediction is commonly posed as scalar regression on a fixed stimulus-locked window, so reaction time is treated as a window-level label rather than timing evidence about response-relevant dynamics. Here we reformulate trial-wise reaction-time decoding as event-time posterior modeling. Instead of predicting a single RT directly, the model estimates a posterior distribution over response-relevant event times, \(p(t_{\mathrm{event}}\mid X)\), and uses the posterior mean as the RT estimate. This treats behavioral latency as a weak observation of latent response-relevant timing, making the output representation itself part of the computational hypothesis.

We evaluate this formulation on the Healthy Brain Network contrast change detection EEG task under a subject-disjoint, release-separated protocol. Across five seeds, distributional event-time supervision consistently improves held-out RT prediction relative to scalar regression and temporal-readout controls. Controlled objective comparisons isolate supervision of the event-time distribution, rather than expectation-based readout alone, as the source of this gain; architecture controls establish that the effect persists across four temporal backbones and is not explained by model scale.

Beyond point prediction, the event-time posterior serves as a diagnostic for separating temporal evidence from scalar shortcuts. Posterior geometry characterizes temporal concentration, target alignment, and interval behavior, while observation-noise calibration separates latent event-time concentration from predictive uncertainty over observed RT. We also use shifted-crop inference to probe shortcut use versus temporal localization. A matched shift-jitter intervention improves shifted-crop robustness, increases mean sensitivity, and moves predictions more often in the expected crop-relative direction. Sensitivity remains below ideal crop-relative localization, leaving a clear equivariance gap. Together, these results establish event-time posterior modeling as a probabilistic and interpretable formulation for linking single-trial EEG dynamics to behavioral timing.

\section{Introduction}

Behavioral timing links neural dynamics to observable behavior. In reaction-time (RT) tasks, the observed response latency reflects sensory encoding, decision formation, motor preparation, and response execution unfolding over hundreds of milliseconds. EEG analyses often exploit this temporal structure through event alignment, response locking, and the study of latency variability. These trial-to-trial timing differences can carry behaviorally meaningful information that may be lost through averaging or when models reduce temporal structure to window-level labels \citep{Jung1998SingleTrialERP,Ouyang2017LatencyReview}.

Supervised EEG decoding, however, often treats a fixed stimulus-locked window as an input to a class label or scalar behavioral variable. For RT decoding, this means that a delayed behavioral response latency is treated as a property of the whole EEG segment. A scalar predictor may therefore achieve low error by exploiting individual differences in typical reaction time, trial difficulty, or stimulus-locked temporal priors, without representing when response-relevant dynamics occur within the window.

We address this mismatch by treating behavioral RT as weak event-time supervision. The button press provides noisy behavioral timing evidence about response-relevant dynamics, reflecting the combined effects of sensory, decision, and motor processes rather than a direct neural event annotation. Given an EEG window \(X\), the model estimates a posterior distribution \(p(t_{\mathrm{event}}\mid X)\) over response-relevant event times and derives the scalar RT prediction as the posterior expectation. This keeps compatibility with standard scalar metrics such as nRMSE while making the timing semantics of RT explicit.

We study this question on the Healthy Brain Network contrast change detection EEG task \citep{Shirazi2024HBNEEG}, following the RT prediction setting introduced by the EEG Foundation Challenge \citep{EEGFoundationChallengeArxiv2025,EEGChallengeWebsite2025}. Under a subject-disjoint, release-separated protocol, we compare direct scalar regression, temporal-readout regression, soft-argmax RT-loss controls, and distributionally supervised event-time models. This design separates the effects of architecture capacity, expectation-based readout, scalar timing loss with a temporal head, and distributional posterior supervision. This output-representation focus complements advances in EEG architectures, self-supervised pretraining, and foundation-style EEG models for encoding EEG across subjects, tasks, and recording setups \citep{wang2024eegpt,Jiang2024LaBraM,Chen2024EEGFormer,Wang2025EEGMamba,ElOuahidi2025REVE,Yue2024BrainGPT}; representative backbones serve as architecture controls within the same evaluation.

The posterior representation also expands what can be evaluated on each trial. Posterior geometry summarizes temporal concentration, target alignment, and interval behavior beyond scalar error, while observation-noise calibration separates latent event-time concentration from predictive uncertainty over observed RT. Shifted-crop inference provides a complementary test of temporal behavior: a crop-relative timing localizer should move its prediction with the temporal frame, whereas a fixed-timing scalar shortcut should remain comparatively invariant.

Distributional event-time supervision consistently improves held-out prediction over scalar and temporal-readout controls across five seeds, and the soft-argmax ablation attributes this advantage to distributional supervision rather than expectation readout alone. Posterior geometry and shifted-crop analyses reveal temporal evidence and partial response-timing localization that scalar metrics do not capture, while shift-jitter training further improves shifted-crop robustness and expected-direction movement. The remaining sensitivity gap provides a clear target for fully equivariant temporal models. Although our experiments focus on RT prediction, the same event-time view is relevant to EEG targets with latency semantics, including ERP latency estimation, movement onset decoding, error-related responses, and other latency-varying phenomena.

Our contributions are:
\begin{enumerate}
    \item We recast behavioral RT as weak event-time supervision and formulate EEG RT decoding as event-time posterior modeling rather than direct scalar regression, with soft-target distribution matching and EventNLL latent-event likelihoods as complementary realizations of this formulation.
    \item We evaluate this formulation under a subject-disjoint, release-separated HBN-EEG protocol and controlled comparison designed to separate EEG backbone capacity, temporal readout parameterization, and distributional event-time supervision.
    \item We show through controlled baselines and loss ablations, repeated across five seeds and four dense temporal backbones, that distributional event-time supervision improves over scalar regression, temporal-readout regression, and soft-argmax RT-loss controls.
    \item We use the learned event-time posterior for temporally resolved evaluation: posterior geometry and observation-noise calibration separate scalar accuracy from temporal concentration, predictive uncertainty, and interval behavior, while shifted-crop diagnostics test shortcut behavior and reveal partial response-timing localization; matched shift-jitter training then strengthens crop-relative behavior.
\end{enumerate}

\section{Related Work}

\subsection{Event-Centered EEG Analysis and Latency Variability}
EEG is often analyzed through events and latencies. Stimulus onset, response execution, error-related activity, movement onset, ERP component timing, and other temporally localized phenomena define not only what occurs in a trial, but also when task-relevant neural or behavioral processes unfold. This event-centered view appears in annotation systems such as Hierarchical Event Descriptors \citep{BigdelyShamlo2016HED} and in single-trial ERP analysis, where ERP-image visualization and related methods show that sorting trials by RT can reveal performance-linked temporal structure \citep{Jung1998SingleTrialERP}. Methods for single-trial ERP analysis further show that within-subject latency variation can be estimated and linked to behavior \citep{Ouyang2011RIDE,Ouyang2017LatencyReview}.

EEG activity associated with responses and movements provides related examples. Lateralized readiness potentials reflect response preparation and execution \citep{Coles1989LRP}. Error-related negativities are response-locked markers of error processing and performance monitoring \citep{Gehring1993ErrorDetection,HolroydColes2002ERN}. Movement-related cortical potentials, including the Bereitschaftspotential, characterize neural activity around movement onset and during premovement preparation \citep{KornhuberDeecke1965BP,ShibasakiHallett2006BP}. Decision-time models make a related point at the behavioral level by treating RT variability as a consequence of latent temporal dynamics in evidence accumulation and decision formation \citep{Ratcliff2008Diffusion,OConnell2012Supramodal,Kelly2013Internal}. At the EEG level, single-trial hidden multivariate pattern analyses make a parallel point by decomposing decision-task EEG into recurrent events with trial-varying latencies \citep{Weindel2025DecisionComponents}. Rather than fitting a cognitive process model or an explicit sequence of neural events, we use observed RT as a behavioral latency that constrains an unobserved response-relevant event time.

Explicit EEG event detection is also well established when annotated onsets, offsets, or intervals are available. Sleep-event detectors, for example, estimate the timing of transient sleep EEG patterns such as sleep spindles and K-complexes rather than predicting only a window-level label \citep{Chambon2019DOSED,TapiaRivas2024SEED}. In our setting, scalar RT serves as weak timing evidence rather than a manually annotated neural event time.

\subsection{Fixed-Window EEG Decoding of Latency-Structured Targets}
Modern supervised EEG decoding often maps a fixed stimulus-locked or task-locked EEG window to a scalar or class label. This formulation is convenient and aligns with benchmark metrics, but it can collapse temporal structure when the target is itself a latency. RT is evaluated as one scalar per trial, yet it denotes the latency of a behavioral response following stimulus onset.

EEG-based RT estimation has commonly followed this scalar-prediction form, including regression pipelines based on Riemannian geometry features \citep{Wu2017RiemannianRT} and deep single-trial EEG models for visual-stimulus RT prediction \citep{Chowdhury2020RTSensors}. The EEG Foundation Challenge provides a standardized scalar benchmark using normalized RMSE \citep{EEGFoundationChallengeArxiv2025,EEGChallengeWebsite2025}. We keep this scalar evaluation protocol, but ask whether the same label can be used within the model as weak event-time supervision. This distinction matters because a scalar regressor may learn trial difficulty, subject-level response tendency, or stimulus-locked temporal priors without representing where response-relevant dynamics are expressed. A temporal soft-argmax readout alone also does not resolve the issue if the model is still trained only through scalar RT error.

\subsection{Distributional and Time-to-Event Output Modeling}
Latency-structured EEG targets also motivate output distributions rather than only point estimates. Label distribution learning replaces a hard label with a distribution over related labels, which is useful when neighboring labels share metric structure or when ambiguity is meaningful \citep{Geng2016LabelDistributionLearning}. For RT, neighboring labels are ordered time bins, making a smoothed event-time distribution a natural supervision target.

Time-to-event modeling provides a complementary probabilistic language for ordered event times. Classical survival analysis and neural time-to-event models estimate distributions over event times rather than only scalar predictions or risk scores \citep{Cox1972CoxPH,Lee2018DeepHit}. Distributional evaluation also makes uncertainty and calibration observable through posterior width, coverage, likelihood scores, and proper scoring rules \citep{Chapfuwa2023Calibration,Haider2020SurvivalDistributions}. We adapt this perspective to finite EEG windows, rather than treating the task as a standard survival-analysis problem with long-horizon risk, censoring, or competing clinical outcomes. EventNLL instantiates this view by treating observed RT as a noisy realization of a latent response-relevant event time.

\subsection{Input Representations, EEG Backbones, and Pretraining}
A separate line of EEG decoding work addresses cross-subject, cross-session, and cross-dataset variability through stronger input representations. Classical approaches include Riemannian covariance-based decoding and alignment methods for reducing subject or session mismatch \citep{Barachant2012RiemannianBCI,Jayaram2016TransferLearningBCI,Zanini2018RiemannianTL,HeWu2020EuclideanAlignment}. Deep learning approaches pursue related goals through domain adaptation, moment matching, supervised EEG architectures, self-supervised pretraining, and foundation-style EEG backbones \citep{Ganin2016DANN,SunSaenko2016DeepCORAL,Kostas2021BENDR,wang2024eegpt,Jiang2024LaBraM,Chen2024EEGFormer,Wang2025EEGMamba,ElOuahidi2025REVE,Yue2024BrainGPT}.

Our contribution advances an output-representation axis that is complementary to these input-representation approaches. This distinction remains important for recent EEG foundation models: they primarily address transferable input representations, whereas our question concerns the output representation appropriate for latency-labeled EEG. Our focus is the representation and supervision of latency-structured outputs: when the target is a latency, the label is not only a scalar value to regress, but also weak evidence about when response-relevant dynamics are expressed. External EEG architectures and foundation-style models trained from scratch therefore serve as controlled architecture baselines. What remains missing in prior work is a controlled EEG RT formulation that connects scalar benchmark evaluation, temporal-readout controls, distributional event-time supervision, and posterior-level diagnostics. We fill this gap by learning a posterior over response-relevant latencies and evaluating it as temporal evidence rather than only as a source of a scalar RT estimate.

\section{Dataset and Evaluation Protocol}
\label{sec:dataset_protocol}

\subsection{HBN-EEG CCD Reaction-Time Task}

We use the reaction-time prediction task introduced by the NeurIPS EEG Foundation Challenge \citep{EEGChallengeWebsite2025,EEGFoundationChallengeArxiv2025}, based on curated releases of the HBN-EEG dataset \citep{Shirazi2024HBNEEG}. Our analysis focuses on the contrast change detection (CCD) reaction-time subset.

In the CCD task, participants viewed two flickering grating stimuli and responded after a contrast change made one grating perceptually dominant. The released target is the trial-wise reaction time from stimulus onset to button press. Each main experiment uses the same fixed 2~s stimulus-locked EEG window spanning 0.5--2.5~s after stimulus onset, sampled at 100~Hz from 128 EEG channels after excluding the reference channel. Thus each input has \(T=200\) samples. Because RT marks the response time within this post-stimulus window, the task admits an event-time interpretation in addition to a window-level regression formulation.

\begin{figure}[!t]
    \centering
    \includegraphics[width=0.9\linewidth]{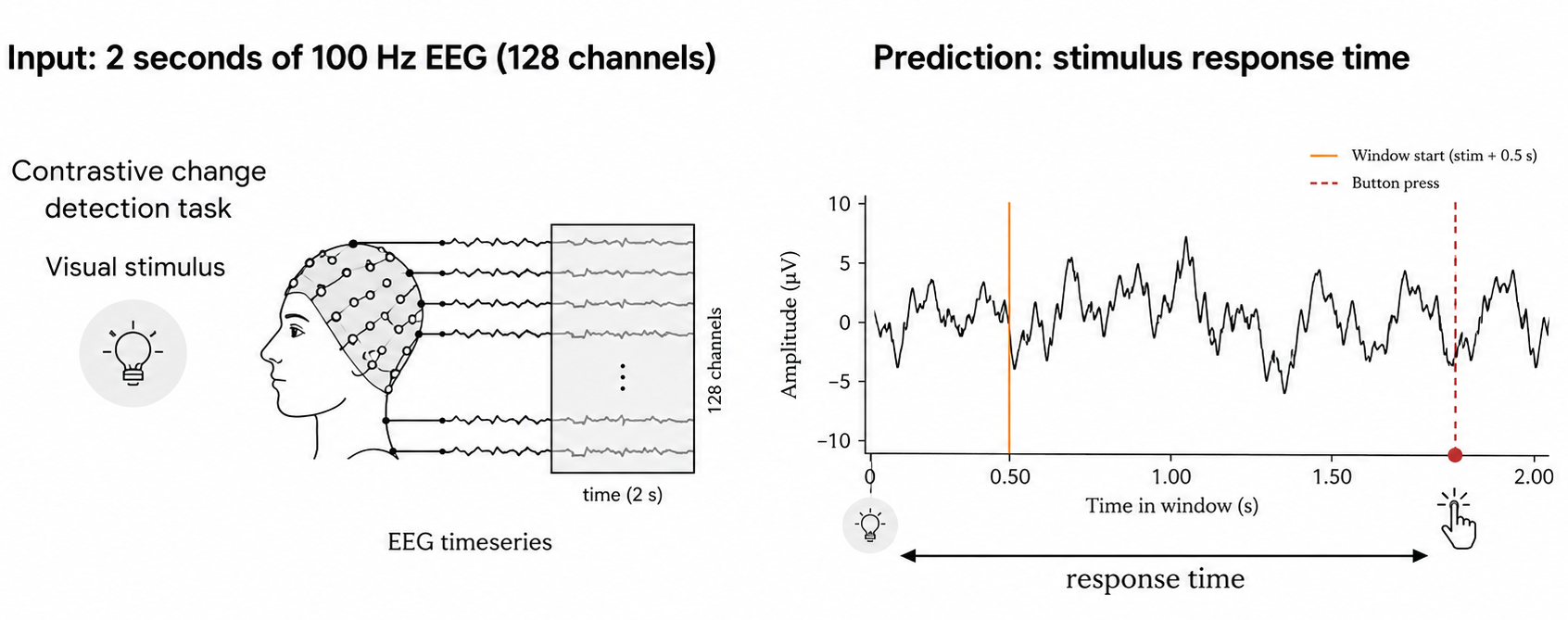}
    \caption{HBN-EEG contrast change detection reaction-time task. Each trial provides a stimulus-locked EEG segment and a trial-wise RT target from stimulus onset to button press.}
    \label{fig:tasks}
\end{figure}

\subsection{Shared Evaluation Protocol}

We use this task as a controlled evaluation setting for separating three possible sources of RT prediction performance: EEG backbone capacity, temporal readout parameterization, and distributional event-time supervision. The benchmark therefore fixes the data split, target support, input representation, and scalar evaluation rule before varying architectures and objectives in the subsequent model and diagnostic sections.

To avoid subject leakage, all models use the same release-separated CCD split: R1--R8 for fitting, R9--R10 as the validation split used for development choices such as early stopping, checkpoint selection, readout-temperature tuning, and diagnostic settings for analyses that require them, and R11 for final holdout evaluation after all model and readout choices are fixed. R11 holdout labels are not used for model fitting, checkpoint selection, readout-temperature selection, or protocol decisions. Table~\ref{tab:ccd_protocol_stats} reports the resulting partitions; metadata checks confirmed zero subject overlap across train, development, and holdout after filtering.

\begin{table}[!t]
\centering
\caption{Release-separated CCD protocol statistics after window construction and RT-support filtering. Prepared trials are 2~s CCD windows with stimulus and response annotations; analyzed trials additionally satisfy \(0.5 \leq \mathrm{RT} \leq 2.5\)~s.}
\label{tab:ccd_protocol_stats}
\setlength{\tabcolsep}{3pt}
\renewcommand{\arraystretch}{1.12}
\begin{tabular*}{\linewidth}{@{\extracolsep{\fill}} l l r r r @{}}
\toprule
\textbf{Partition} & \textbf{Releases} & \textbf{Prepared trials} & \textbf{Analyzed trials} & \textbf{Subjects} \\
\midrule
Train & R1--R8 & 74{,}576 & 73{,}030 & 1{,}221 \\
Development & R9--R10 & 17{,}867 & 17{,}348 & 308 \\
Holdout & R11 & 15{,}751 & 15{,}164 & 292 \\
\bottomrule
\end{tabular*}
\end{table}

The main analysis keeps trials whose behavioral RT falls inside the modeled event-time support, 0.5--2.5~s after stimulus onset. This filter removes 1{,}546/74{,}576 training trials (2.07\%), 519/17{,}867 development trials (2.90\%), and 587/15{,}751 holdout trials (3.73\%). In the prepared CCD split, excluded trials are fast responses below 0.5~s; no prepared trial has RT above 2.5~s. The conclusions therefore apply to RTs within the modeled 0.5--2.5~s post-stimulus interval.

Following the EEG Foundation Challenge evaluation protocol \citep{EEGChallengeWebsite2025}, scalar performance is evaluated using normalized RMSE (nRMSE), computed as RMSE divided by the standard deviation of the true targets within each evaluated split. Main scalar comparisons use the same support-filtered R9--R10 validation set and final holdout set. Diagnostic analyses that require shifted crop starts, stricter RT support, or posterior-only summaries are introduced in the corresponding subsection or appendix and do not affect the main holdout reporting protocol.

\begin{table}[!t]
\centering
\caption{Controlled comparison and diagnostic blocks under the shared protocol.}
\label{tab:experiment_map}
\scriptsize
\setlength{\tabcolsep}{2pt}
\renewcommand{\arraystretch}{1.08}
\begin{tabularx}{\linewidth}{@{} >{\RaggedRight\arraybackslash}p{0.21\linewidth} >{\RaggedRight\arraybackslash}p{0.18\linewidth} >{\RaggedRight\arraybackslash}p{0.18\linewidth} >{\RaggedRight\arraybackslash}X @{}}
\toprule
\textbf{Control or analysis} & \textbf{Readout} & \textbf{Supervision} & \textbf{Question addressed} \\
\midrule
\multicolumn{4}{@{}l}{\textit{Scalar regression controls}} \\
\addlinespace[0.15em]
External EEG backbones & Scalar RT head & Scalar RT & Does generic EEG backbone capacity explain performance? \\
MSP-CNN family & Segment-pooling scalar readout & Scalar RT & How far does coarse temporal pooling support scalar regression? \\
ETR-CNN family & Temporal expectation readout & Scalar RT & Does a learned temporal readout improve scalar regression? \\
\midrule
\multicolumn{4}{@{}l}{\textit{Event-time objective comparison with fixed ETS-U-Net}} \\
\addlinespace[0.15em]
Soft-argmax RT-loss control & Posterior mean & Scalar RT only & Is posterior-mean readout sufficient without distributional supervision? \\
CE soft-target objective & Posterior mean & Soft event-time target & Does direct distributional supervision improve RT prediction? \\
Likelihood-based objectives & Posterior mean & Latent event-time likelihood & Does probabilistic latent-event supervision provide the same benefit? \\
Wasserstein control & Posterior mean & Distributional geometry loss & Is an alternative geometry-based distributional objective sufficient? \\
\midrule
\multicolumn{4}{@{}l}{\textit{Architecture robustness controls}} \\
\addlinespace[0.15em]
ETS-TCN & Posterior mean & RT-only, CE, mixture EventNLL & Does the supervision effect persist with dilated temporal convolution? \\
ETS-InceptionPyramid & Posterior mean & RT-only, CE, mixture EventNLL & Does the supervision effect persist with multi-scale temporal filtering? \\
ETS-AttnSeg & Posterior mean & RT-only, CE, mixture EventNLL & Does the supervision effect persist with attention and local convolution? \\
\midrule
\multicolumn{4}{@{}l}{\textit{Posterior and localization diagnostics}} \\
\addlinespace[0.15em]
Readout-temperature tuning & Temperature-adjusted posterior mean & Learned posterior & How should the scalar posterior readout be standardized across objectives? \\
Posterior geometry diagnostics & Posterior summaries & Learned posterior & What posterior behavior is hidden by scalar nRMSE? \\
Shifted-crop diagnostic & Crop-relative posterior mean & No new training & Do predictions behave as crop-relative temporal localizers? \\
Shift-jitter intervention & Crop-relative posterior mean & Shifted crop-relative targets & Does crop-start augmentation reduce shortcut behavior? \\
\bottomrule
\end{tabularx}
\end{table}

Table~\ref{tab:experiment_map} maps the shared protocol to the comparison blocks used below. The scalar regression controls vary compact temporal regressors and external EEG backbones under scalar RT supervision. We first fix ETS-U-Net and vary output supervision (Section~\ref{sec:event_time_architectures}), and then repeat a reduced objective set across additional backbones to test architecture robustness (Section~\ref{sec:architecture_robustness}). Posterior geometry and shifted-crop inference are post-training diagnostics, while shift-jitter training is a matched intervention; together they evaluate posterior semantics and shortcut-vs-localization behavior within the same benchmark framework.

\subsection{Shared Training, Readout Tuning, and Inference Protocol}
\label{sec:shared_protocol}

To support the controlled comparison, we keep preprocessing, optimization, augmentation, checkpointing, and inference fixed across supervised neural models unless stated otherwise. Each 2~s trial window is normalized per channel by subtracting the channel's temporal mean within the window and dividing by its temporal standard deviation. The shared optimization and augmentation settings are summarized in Appendix~\ref{app:reproducibility_details}; in brief, models are trained with Adam \citep{kingma2015adam}, early stopping on validation nRMSE, and a fixed augmentation profile combining channel dropout, temporal cutout, and Gaussian noise.

Regression models use scalar RT supervision, while event-time models represent RT either as a soft crop-relative event-time target or as an observation under a latent event-time likelihood. When temperature scaling is used for event-time posterior readout, \(\tau\) is selected on R9--R10 to minimize development nRMSE of the posterior-mean readout and then applied unchanged to the holdout split. The main regression-baseline and event-time-objective comparisons are repeated over five seeds (2025--2029), and repeated-run tables report mean $\pm$ standard deviation across seeds unless otherwise stated.

\section{Scalar Regression Controls}
\label{sec:baseline_regression}

Before introducing event-time targets, we first ask how far scalar RT supervision can go under the same release-separated protocol. The regression baselines use compact task-specific temporal regressors as strong scalar controls and external EEG backbones as architecture-capacity controls.

\begin{figure}[!t]
    \centering
    \includegraphics[width=\linewidth]{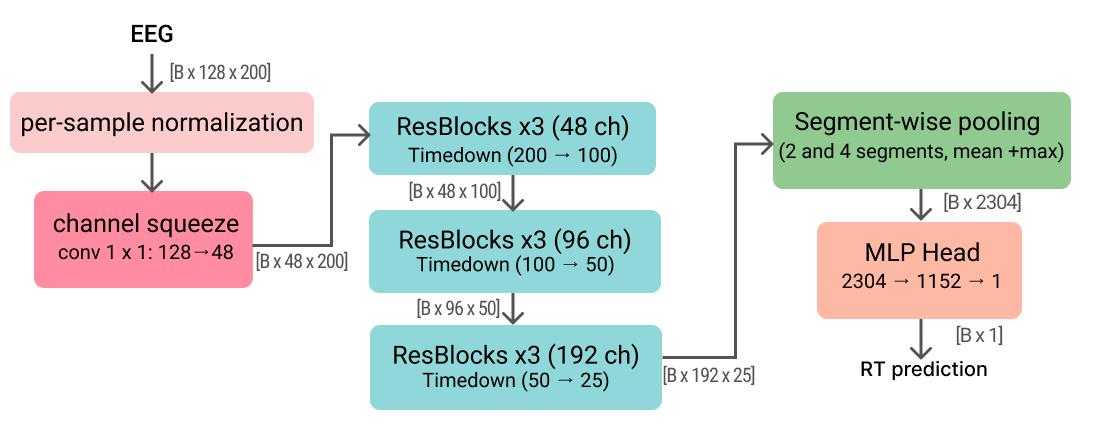}
    \caption{MSP-CNN direct RT regression baseline architecture. A lightweight 1-D temporal ConvNet maps an EEG trial window to a scalar reaction-time prediction. A learned $1\times1$ channel-squeeze layer (128$\rightarrow$48) is followed by residual depthwise-separable stages with anti-aliased temporal downsampling, segment-wise pooling (2 and 4 segments; mean+max), and an MLP head.}
    \label{fig:ch1_regression}
\end{figure}

The compact controls form a progression of window-level temporal inductive biases. The multiscale segment-statistic pooling CNN (MSP-CNN; Figure~\ref{fig:ch1_regression}) explicitly accounts for temporal structure by collecting mean and max statistics from coarse temporal segments of the feature map. These segment-level summaries act as early/middle/late activation cues and are mapped to scalar RT. The event-time readout CNN (ETR-CNN) instead produces learned temporal scores over the window and predicts RT as their expectation, while training the resulting scalar prediction through RT error. ETR-CNN large keeps the same readout while increasing feature capacity.

We compare these task-specific regressors with external EEG architectures adapted to scalar RT regression. These include classical convolutional EEG baselines, Deep4Net and ShallowFBCSPNet \citep{braindecode}; the compact EEGNet architecture \citep{lawhern2018eegnet}; TIDNet's thinker-invariant convolutional design \citep{kostas2020thinkerinvariance}; the convolutional-transformer EEGConformer \citep{song2023eegconformer}; the attention temporal convolutional ATCNet architecture \citep{altaheri2022atcnet}; EEGPT \citep{wang2024eegpt}; Medformer \citep{wang2024medformer}; and the LaBraM foundation-style architecture \citep{Jiang2024LaBraM}. All external architectures are trained from scratch; pretrained weights are not used.

ETR-CNN large is the strongest scalar baseline, with MSP-CNN and base ETR-CNN close behind. Under this protocol, the evaluated external EEG backbones remain below the task-specific temporal controls, indicating that generic backbone capacity alone does not account for the stronger scalar performance. These results establish ETR-CNN large as the scalar reference for testing whether direct event-time distribution supervision improves beyond scalar RT training.

\begin{table}[!t]
\centering
\caption{Direct-regression and external backbone baselines.}
\label{tab:r11_regression_baselines}
\footnotesize
\setlength{\tabcolsep}{2pt}
\renewcommand{\arraystretch}{1.15}
\begin{tabularx}{\linewidth}{@{} >{\RaggedRight\arraybackslash}p{0.22\linewidth} >{\RaggedRight\arraybackslash}X >{\centering\arraybackslash}p{0.215\linewidth} >{\centering\arraybackslash}p{0.215\linewidth} @{}}
\toprule
\textbf{Model} & \textbf{Role} & \textbf{Valid nRMSE} $\downarrow$ & \textbf{Holdout nRMSE} $\downarrow$ \\
\midrule
\multicolumn{4}{@{}l}{\textit{Task-specific regression baselines}} \\
\addlinespace[0.15em]
MSP-CNN & segment-pooling scalar & \mbox{0.9006 $\pm$ 0.0051} & \mbox{0.8998 $\pm$ 0.0080} \\
ETR-CNN & temporal readout & \mbox{0.9008 $\pm$ 0.0060} & \mbox{0.8977 $\pm$ 0.0068} \\
ETR-CNN large & capacity ablation & \textbf{\mbox{0.8972 $\pm$ 0.0040}} & \textbf{\mbox{0.8928 $\pm$ 0.0042}} \\
\midrule
\multicolumn{4}{@{}l}{\textit{External regression baselines}} \\
\addlinespace[0.15em]
TIDNet & thinker-invariant CNN & \mbox{0.9235 $\pm$ 0.0024} & \textbf{\mbox{0.9192 $\pm$ 0.0027}} \\
EEGConformer & conv-transformer & \textbf{\mbox{0.9188 $\pm$ 0.0026}} & \mbox{0.9287 $\pm$ 0.0057} \\
EEGNet & compact CNN & \mbox{0.9350 $\pm$ 0.0054} & \mbox{0.9335 $\pm$ 0.0028} \\
LaBraM & foundation-style & \mbox{0.9304 $\pm$ 0.0061} & \mbox{0.9327 $\pm$ 0.0086} \\
Deep4Net & deep ConvNet & \mbox{0.9269 $\pm$ 0.0045} & \mbox{0.9260 $\pm$ 0.0044} \\
ShallowFBCSPNet & shallow FBCSP CNN & \mbox{0.9324 $\pm$ 0.0016} & \mbox{0.9343 $\pm$ 0.0024} \\
ATCNet & conv/attention/TCN & \mbox{0.9686 $\pm$ 0.0183} & \mbox{0.9666 $\pm$ 0.0151} \\
EEGPT & foundation-style & \mbox{0.9616 $\pm$ 0.0201} & \mbox{0.9584 $\pm$ 0.0185} \\
Medformer & larger transformer & \mbox{0.9623 $\pm$ 0.0046} & \mbox{0.9585 $\pm$ 0.0051} \\
\bottomrule
\end{tabularx}
\end{table}

\section{Event-Time Posterior Formulation}
\label{sec:event_time_formulation}

To make response timing explicit, we reformulate RT prediction as event-time posterior modeling. Instead of predicting a scalar directly, the model produces per-time logits and a posterior distribution over response-relevant event times; scalar RT is then read out as the posterior expectation. The key distinction from temporal-readout regression is that the distributionally supervised objectives constrain the event-time representation itself, while the RT-only control uses the same posterior-mean readout with scalar supervision.

We evaluate two ways of converting scalar RT labels into event-time supervision. The first constructs an explicit soft target distribution over time, asking the model to match a smoothed target centered at the observed RT. The second treats the observed RT statistically, as a noisy observation of an unobserved response-relevant event time, and optimizes a latent event-time likelihood. These two families correspond to soft-target objectives and likelihood-based objectives, respectively.

The core event-time experiments are organized as a primary controlled comparison
and a secondary architecture-robustness analysis. The regression controls above quantify
what scalar supervision achieves with task-specific temporal readouts and external EEG
backbones.

We use the event-time segmentation U-Net (ETS-U-Net) as the primary backbone
for the objective comparisons because its dense temporal segmentation design
naturally matches the event-time output space: it maps each EEG window to
per-time logits while combining local and broader temporal context through an
encoder--decoder path with skip connections. Fixing this backbone allows us to
isolate the effect of output supervision under a strong segmentation model.
Additional dense temporal backbones then test whether the resulting supervision
effects persist beyond the U-Net inductive bias, as described in
Section~\ref{sec:architecture_robustness}.

\begin{figure}[!t]
    \centering
    \includegraphics[width=\linewidth]{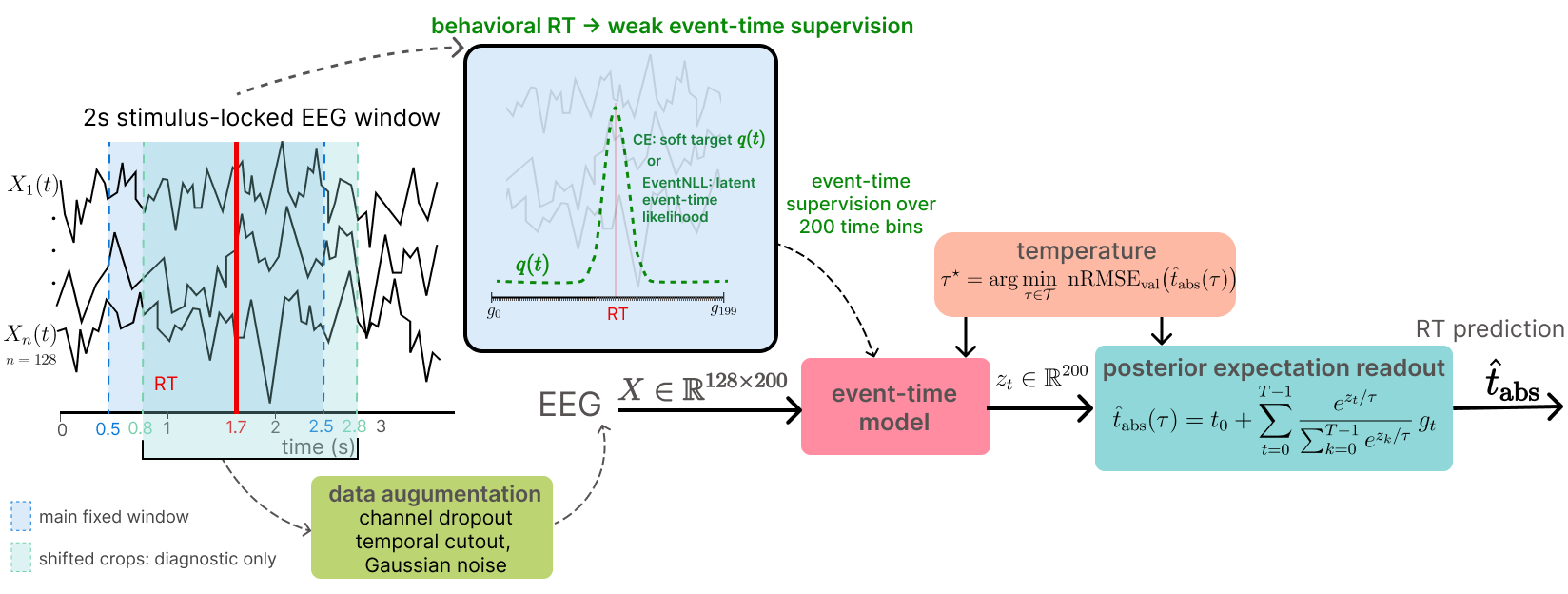}
    \caption{Event-time posterior formulation for EEG reaction-time prediction. A fixed 0.5--2.5~s stimulus-locked EEG window is mapped to time-bin logits, and behavioral RT provides weak event-time supervision either through a soft target distribution or a latent EventNLL likelihood. Scalar RT is read out as the posterior expectation after selecting a readout temperature on R9--R10. Shifted crops are used only for diagnostic and shift-jitter intervention analyses. The schematic shows the categorical posterior readout; the hazard EventNLL variant uses a hazard-derived probability mass function with the same expectation readout.}
    \label{fig:event_time_scheme}
\end{figure}

\subsection{Primary Event-Time Segmentation Architecture}
\label{sec:event_time_architectures}

\begin{figure}[t]
    \centering
    \includegraphics[width=0.9\linewidth]{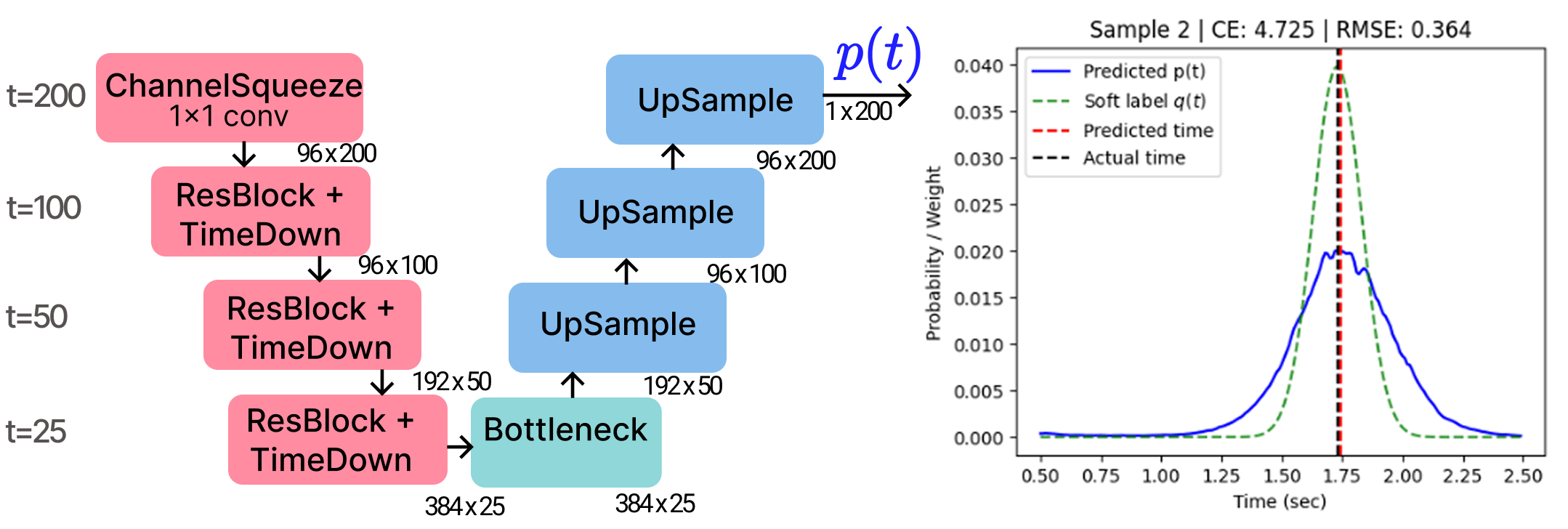}
    \caption{ETS-U-Net for RT segmentation. A lightweight 1-D U-Net maps an EEG trial window to per-time logits
    and an event-time distribution $p(t)$. The encoder applies a $1\times1$ channel-squeeze layer followed by residual
    blocks with temporal downsampling (\texttt{TimeDown}); a bottleneck aggregates context, and the decoder upsamples with
    skip connections to recover time-resolved outputs. The right panel shows an example prediction $p(t)$ and Gaussian soft
    target $q(t)$, together with the resulting predicted vs.\ observed response time.}
    \label{fig:ets_unet}
\end{figure}

ETS-U-Net maps an EEG trial window
$X\in\mathbb{R}^{C\times T}$ to per-time logits $z_{0:T-1}$ (Figure~\ref{fig:ets_unet}). It follows a 1-D U-Net style encoder-decoder with skip
connections that preserve fine temporal resolution \citep{ronneberger2015unet}. The encoder progressively downsamples the temporal axis
to build multi-scale context, while the decoder upsamples and fuses encoder features to recover per-time outputs at the
input temporal resolution (downsampling and upsampling are used only to form multi-scale features). Each scale is refined
with lightweight residual blocks (ResNet-style skips) \citep{he2016resnet} built from depthwise-separable 1-D convolutions
for computational efficiency \citep{howard2017mobilenets,chollet2017xception}. To mitigate aliasing under temporal
decimation, downsampling applies a fixed depthwise smoothing filter followed by strided averaging (``blur-pool'' style)
\citep{zhang2019shiftinvariant}. A dilated bottleneck further enlarges the receptive field without additional downsampling
\citep{yu2016dilated}.

\subsection{Soft-Target Distribution Matching}

The explicit-target formulation begins by placing each observed RT \(y\) on the
model's event-time grid. Let \(\Delta t=10\)~ms denote the grid spacing, and let
\(g_t=t\Delta t\), \(t=0,\ldots,T-1\), denote the crop-relative time assigned to
bin \(t\). We convert the observed RT from stimulus-relative to crop-relative
coordinates, \(y_{\mathrm{rel}}=y-t_0\), where $t_0$ is the crop start, and
construct a Gaussian soft target with bandwidth $\sigma$:
\begin{equation}
\label{eq:rt_soft_target}
\tilde q_t(y)
=e^{-\frac{(g_t-y_{\mathrm{rel}})^2}{2\sigma^2}},
\qquad
q_t(y)
=\frac{\tilde q_t(y)}{\sum_{k=0}^{T-1}\tilde q_k(y)},
\qquad
\sum_{t=0}^{T-1}q_t(y)=1.
\end{equation}
The resulting Gaussian target $q_y=(q_0(y),\ldots,q_{T-1}(y))$ provides local
temporal smoothing around the observed latency; its bandwidth $\sigma$ is a
supervision parameter rather than an estimate of RT uncertainty.

Temporal logits $z_{\theta,t}(X)$ define the training posterior
\begin{equation}
\label{eq:rt_pred_dist}
p_{\theta,t}(X)
=\frac{e^{z_{\theta,t}(X)/\tau_0}}
{\sum_{k=0}^{T-1}e^{z_{\theta,k}(X)/\tau_0}},
\end{equation}
where $\tau_0$ is fixed during training.

\paragraph{Distribution-matching comparison.}
We first ask how the predicted posterior should be matched to the constructed
target. In general, this objective can be written as
\begin{equation}
\label{eq:rt_loss}
\mathcal{L}_{D}(X,y)
=D\!\left(q_y,p_\theta(\cdot\mid X)\right),
\end{equation}
where $D$ determines how disagreement between the two distributions is
penalized. We use cross-entropy (CE) as the primary soft-target objective and the
one-dimensional Wasserstein distance $W_1$ as a geometry-aware control:
\begin{equation}
D=\mathrm{CE}
\quad\Longrightarrow\quad
\mathcal{L}_{\mathrm{CE}}(X,y)
=-\sum_{t=0}^{T-1}q_t(y)\log p_{\theta,t}(X).
\end{equation}
\begin{equation}
D=W_1
\quad\Longrightarrow\quad
\mathcal{L}_{W_1}(X,y)
=\Delta t\sum_{t=0}^{T-1}
\left|\mathrm{CDF}(q_y)_t
-\mathrm{CDF}\!\left(p_\theta(\cdot\mid X)\right)_t\right|.
\end{equation}
Cross-entropy treats $q_y$ as a soft categorical target, with temporal proximity
represented through the Gaussian smoothing of the target itself. By contrast,
$W_1$ explicitly uses the ordering of the temporal grid and measures how far
probability mass must move to match the target.

\paragraph{Posterior-mean control.}
To test whether distributional supervision adds beyond the posterior-mean
readout, we use an RT-only control:
\begin{equation}
\hat t_{\mathrm{rel}}(X)
=\sum_{t=0}^{T-1}g_t\,p_{\theta,t}(X),\qquad
\mathcal{L}_{\mathrm{RT}}
=\mathrm{RMSE}\!\left(\hat t_{\mathrm{rel}}(X),y_{\mathrm{rel}}\right).
\end{equation}
CE and $W_1$ supervise the full event-time distribution, whereas the RT-only
control supervises only the scalar prediction produced by the posterior-mean
readout.

\subsection{Likelihood-Based Event-Time Objectives}
\label{sec:event_nll}

As a probabilistic alternative to soft-target supervision, the likelihood-based
formulation uses the same event-time posterior but does not construct an explicit
target distribution $q_y$. Instead, it treats the observed RT $y$ as a noisy
observation of a latent response-relevant event at stimulus-relative time
$t_0+g_t$. Given an observation kernel $K$, the marginal likelihood is
\begin{equation}
\label{eq:event_nll}
p_\theta(y\mid X)
=\sum_{t=0}^{T-1}p_{\theta,t}(X)\,
K\!\left(y\mid t_0+g_t\right),
\qquad
\mathcal{L}_{\mathrm{EventNLL}}(X,y)
=-\log p_\theta(y\mid X).
\end{equation}
Here, $p_{\theta,t}(X)$ represents the posterior over latent event times, whereas
$K$ describes how a latent event time generates the observed behavioral RT. This
separation is also consistent with measurement-oriented views of learned
representations as proxy measurements of latent variables
\citep{Yao2025ThirdPillar}.

\paragraph{Observation-kernel comparison.}
We first vary the observation kernel while keeping the posterior parameterization
and marginal-likelihood objective fixed. The default model uses a Gaussian kernel,
\begin{equation}
K_{\mathrm{G}}(y\mid t)=\mathcal{N}(y;t,\sigma_y^2).
\end{equation}
We also evaluate a two-scale Gaussian mixture,
\begin{equation}
\label{eq:event_nll_mixture_kernel}
K_{\mathrm{mix}}(y\mid t)
=(1-\alpha)\,\mathcal{N}(y;t,\sigma_{\mathrm{narrow}}^2)
+\alpha\,\mathcal{N}(y;t,\sigma_{\mathrm{wide}}^2),
\end{equation}
which keeps most probability mass near the latent event time while allowing a
wider component for occasional noisy RT observations. The observation-kernel
parameters are reported in Table~\ref{tab:reproducibility_details}.

\paragraph{Posterior-parameterization comparison.}
Motivated by discrete-time survival modeling, we also evaluate a hazard-based
construction of the event-time posterior \citep{Gensheimer2019NnetSurvival}.
Whereas a categorical softmax assigns probability to all time bins through a
single global normalization, $h_{\theta,t}(X)$ represents the conditional
probability that the event occurs in bin $t$, given that it has not occurred
earlier. The corresponding event-time PMF combines this current-bin probability
with the probability of reaching bin $t$ without an earlier event:
\begin{equation}
\label{eq:hazard_readout}
h_{\theta,t}(X)
=\operatorname{sigmoid}\!\left(z_{\theta,t}(X)/\tau_0\right),\qquad
p_{\theta,t}^{\mathrm{haz}}(X)
\propto h_{\theta,t}(X)\prod_{k<t}\left(1-h_{\theta,k}(X)\right),
\end{equation}
The normalized $p_{\theta,t}^{\mathrm{haz}}(X)$ replaces $p_{\theta,t}(X)$ in
Eq.~\ref{eq:event_nll}.

\subsection{Formulation Comparison and Robustness}

We next compare the event-time objectives under a fixed ETS-U-Net backbone. The central question is whether distributional event-time supervision improves RT prediction beyond posterior-mean readout alone, and whether this benefit depends on a particular target construction, observation kernel, or posterior parameterization. Table~\ref{tab:r11_segmentation_losses} reports all variants under the same release-separated protocol. Holdout \(\tau\)-nRMSE uses the readout temperature selected on R9--R10 and applied unchanged to the holdout split. After this common readout procedure, CE and the likelihood-based variants achieve the lowest held-out errors, with holdout \(\tau\)-nRMSE between 0.8745 and 0.8778. These objectives outperform the strongest scalar regression baseline, ETR-CNN large (0.8928 $\pm$ 0.0042; Table~\ref{tab:r11_regression_baselines}), whereas the soft-argmax RT-loss and Wasserstein controls remain closer to the scalar-regression baselines.

ETR-CNN large provides a stringent scalar reference because it already includes a task-specific temporal expectation readout. Event-time supervision reduces held-out error beyond this scalar temporal inductive bias.

\begin{table}[!htbp]
\centering
\caption{Event-time objective variants with a fixed ETS-U-Net backbone.}
\label{tab:r11_segmentation_losses}
\footnotesize
\setlength{\tabcolsep}{2pt}
\renewcommand{\arraystretch}{1.15}
\begin{tabularx}{\linewidth}{@{} >{\RaggedRight\arraybackslash}p{0.18\linewidth} >{\RaggedRight\arraybackslash}X >{\centering\arraybackslash}p{0.17\linewidth} >{\centering\arraybackslash}p{0.17\linewidth} >{\centering\arraybackslash}p{0.17\linewidth} @{}}
\toprule
\textbf{Objective} & \textbf{Supervision signal} & \textbf{Valid nRMSE} & \textbf{Holdout nRMSE} & \textbf{Holdout \(\tau\)-nRMSE} \\
\midrule
\multicolumn{5}{@{}l}{\textit{Soft-target objectives}} \\
\addlinespace[0.15em]
CE & Gaussian soft event-time label & \mbox{0.8763 $\pm$ 0.0044} & \textbf{\mbox{0.8774 $\pm$ 0.0044}} & \mbox{0.8753 $\pm$ 0.0039} \\
Wasserstein & CDF match to Gaussian soft event-time label & \mbox{0.8997 $\pm$ 0.0035} & \mbox{0.8995 $\pm$ 0.0078} & \mbox{0.8896 $\pm$ 0.0033} \\
\midrule
\multicolumn{5}{@{}l}{\textit{Likelihood-based objectives}} \\
\addlinespace[0.15em]
EventNLL & latent event time + Gaussian RT likelihood & \mbox{0.8769 $\pm$ 0.0030} & \mbox{0.8805 $\pm$ 0.0021} & \mbox{0.8772 $\pm$ 0.0018} \\
Mixture EventNLL & Two-component core-tail RT likelihood & \textbf{\mbox{0.8744 $\pm$ 0.0018}} & \mbox{0.8785 $\pm$ 0.0047} & \textbf{\mbox{0.8745 $\pm$ 0.0053}} \\
Hazard EventNLL & hazard posterior + Gaussian RT likelihood & \mbox{0.8755 $\pm$ 0.0027} & \mbox{0.8776 $\pm$ 0.0031} & \mbox{0.8778 $\pm$ 0.0041} \\
\midrule
\multicolumn{5}{@{}l}{\textit{Readout-only control}} \\
\addlinespace[0.15em]
Soft-argmax RT loss & posterior-mean RT error only; no distribution target & \mbox{0.8944 $\pm$ 0.0048} & \mbox{0.8943 $\pm$ 0.0025} & \mbox{0.8917 $\pm$ 0.0046} \\
\bottomrule
\end{tabularx}
\end{table}
\FloatBarrier

\paragraph{Objective ablations.}
The ablations separate expectation-style temporal readout from distributional supervision. The soft-argmax RT-loss control directly optimizes posterior-mean RT error but removes distribution matching; it remains close to the strongest scalar-regression baseline and trails the CE/EventNLL group by roughly 0.014--0.017 holdout \(\tau\)-nRMSE. CE and the likelihood-family variants form a tight cluster among the best-performing objectives: mixture EventNLL has the best mean temperature-tuned scalar readout, but CE, Gaussian EventNLL, mixture EventNLL, and hazard EventNLL are close relative to seed variability. The hazard result shows that the likelihood-family conclusion is not tied to a categorical softmax posterior, while the Wasserstein control shows that an alternative distributional distance alone does not match the CE/EventNLL scalar readouts. Within the fixed ETS-U-Net backbone, these results isolate the gain to the supervision objective rather than to the temporal expectation readout alone. The next subsection tests whether the same supervision pattern persists when the dense temporal backbone is changed.

\paragraph{Practical effect size.}
To express the main accuracy gain on the RT scale, we compared the strongest scalar baseline, ETR-CNN large, with the strongest event-time objectives using matched seeds and subject-level bootstrap intervals on the holdout split. Mixture EventNLL reduced \(\tau\)-nRMSE from \(0.8928 \pm 0.0042\) to \(0.8745 \pm 0.0053\), an absolute reduction of \(0.0184\), or about \(2.1\%\) relative improvement. On the RT scale, this corresponds to a \(6.24\)~ms reduction in RMSE and an \(8.94\)~ms reduction in MAE, with subject-bootstrap 95\% confidence intervals of \([4.50, 7.97]\)~ms and \([7.43, 10.49]\)~ms, respectively. The improvement was positive in all five matched-seed comparisons, and both bootstrap intervals exclude zero. Together, the matched-seed and subject-bootstrap results establish a consistent predictive advantage over the scalar temporal-readout reference. The event-time formulation also provides a common posterior object for the diagnostics and shifted-crop analyses in Section~\ref{sec:posterior_diagnostics}.

\subsection{Architecture Robustness of the Supervision Effect}
\label{sec:architecture_robustness}

To assess whether the supervision effect is robust to backbone choice, we
repeat the RT-only soft-argmax, CE, and mixture EventNLL comparison using three
custom temporal backbones. These controls preserve full-resolution event-time
output, closely match ETS-U-Net in capacity, and span three distinct temporal
inductive biases.

ETS-TCN uses residual dilated temporal convolutions
\citep{bai2018tcn,yu2016dilated}. ETS-InceptionPyramid applies parallel
depthwise temporal filters at multiple receptive-field scales and fuses them
through residual blocks, following Inception-style design principles
\citep{szegedy2015going,santamaria2020eeginception}. ETS-AttnSeg uses
Conformer-style blocks that combine global temporal self-attention with local
gated depthwise convolution
\citep{vaswani2017attention,gulati2020conformer}. The three controls contain
3.05--3.25 million trainable parameters, compared with 3.10 million for
ETS-U-Net. Table~\ref{tab:architecture_robustness} reports the resulting scalar RT
comparison.

\begin{table}[!htbp]
\centering
\caption{Scalar RT accuracy across event-time backbones and objectives.}
\label{tab:architecture_robustness}
\footnotesize
\setlength{\tabcolsep}{3pt}
\renewcommand{\arraystretch}{1.15}
\begin{tabularx}{\linewidth}{@{} >{\RaggedRight\arraybackslash}X >{\centering\arraybackslash}p{0.22\linewidth} >{\centering\arraybackslash}p{0.22\linewidth} >{\centering\arraybackslash}p{0.22\linewidth} @{}}
\toprule
\textbf{Objective} & \textbf{Valid nRMSE} & \textbf{Holdout nRMSE} & \textbf{Holdout $\tau$-nRMSE} \\
\midrule
\multicolumn{4}{@{}l}{\textit{ETS-U-Net}} \\
\addlinespace[0.15em]
RT-only soft-argmax & \mbox{0.8944 $\pm$ 0.0048} & \mbox{0.8943 $\pm$ 0.0025} & \mbox{0.8917 $\pm$ 0.0046} \\
CE & \mbox{0.8763 $\pm$ 0.0044} & \mbox{0.8774 $\pm$ 0.0044} & \mbox{0.8753 $\pm$ 0.0039} \\
Mixture EventNLL & \mbox{0.8744 $\pm$ 0.0018} & \mbox{0.8785 $\pm$ 0.0047} & \textbf{\mbox{0.8745 $\pm$ 0.0053}} \\
\midrule
\multicolumn{4}{@{}l}{\textit{ETS-TCN}} \\
\addlinespace[0.15em]
RT-only soft-argmax & \mbox{0.8865 $\pm$ 0.0029} & \mbox{0.8853 $\pm$ 0.0044} & \mbox{0.8842 $\pm$ 0.0045} \\
CE & \mbox{0.8717 $\pm$ 0.0025} & \mbox{0.8751 $\pm$ 0.0085} & \mbox{0.8730 $\pm$ 0.0038} \\
Mixture EventNLL & \mbox{0.8718 $\pm$ 0.0046} & \mbox{0.8729 $\pm$ 0.0020} & \textbf{\mbox{0.8722 $\pm$ 0.0021}} \\
\midrule
\multicolumn{4}{@{}l}{\textit{ETS-InceptionPyramid}} \\
\addlinespace[0.15em]
RT-only soft-argmax & \mbox{0.8970 $\pm$ 0.0043} & \mbox{0.8894 $\pm$ 0.0054} & \mbox{0.8886 $\pm$ 0.0044} \\
CE & \mbox{0.8710 $\pm$ 0.0037} & \mbox{0.8717 $\pm$ 0.0017} & \textbf{\mbox{0.8717 $\pm$ 0.0036}} \\
Mixture EventNLL & \mbox{0.8703 $\pm$ 0.0018} & \mbox{0.8780 $\pm$ 0.0026} & \mbox{0.8746 $\pm$ 0.0028} \\
\midrule
\multicolumn{4}{@{}l}{\textit{ETS-AttnSeg}} \\
\addlinespace[0.15em]
RT-only soft-argmax & \mbox{0.9114 $\pm$ 0.0350} & \mbox{0.9105 $\pm$ 0.0294} & \mbox{0.9052 $\pm$ 0.0300} \\
CE & \mbox{0.8742 $\pm$ 0.0036} & \mbox{0.8811 $\pm$ 0.0097} & \mbox{0.8780 $\pm$ 0.0082} \\
Mixture EventNLL & \mbox{0.8719 $\pm$ 0.0082} & \mbox{0.8823 $\pm$ 0.0079} & \textbf{\mbox{0.8752 $\pm$ 0.0050}} \\
\bottomrule
\end{tabularx}
\end{table}

Across all four backbones, both CE and mixture EventNLL yield lower mean
holdout $\tau$-nRMSE than the RT-only soft-argmax control. Although the leading
distributional objective varies between CE and mixture EventNLL, this
supervision advantage replicates across U-Net, dilated-convolution,
multi-scale convolution, and attention-convolution designs. Complementary
shifted-crop and posterior-geometry analyses are reported in
Appendix~\ref{app:architecture_controls}.
\FloatBarrier

\section{Posterior Readout and Diagnostics}
\label{sec:posterior_diagnostics}

\subsection{Scalar Readout and Temperature Tuning}

\paragraph{Posterior-mean scalar readout.}
Event-time models output a posterior over response-relevant time bins, whereas the benchmark evaluates scalar RT. We therefore use the posterior expectation as the scalar readout:
\begin{equation}
\label{eq:rt_soft_argmax}
\hat t_{\mathrm{rel}}(\tau)=\sum_{t=0}^{T-1} p_t(\tau)\, g_t,
\qquad
\hat t_{\mathrm{abs}}(\tau)=t_0+\hat t_{\mathrm{rel}}(\tau),
\end{equation}
where $g_t=t\,\Delta t$ denotes the discrete time grid. The relative prediction is converted to absolute RT by adding the fixed window start $t_0$, keeping event-time models comparable to scalar regression baselines under nRMSE.

\paragraph{Readout-temperature tuning.}
Because different objectives can produce posteriors with different sharpness, we select a readout temperature on the R9--R10 development split to minimize posterior-mean nRMSE and apply it unchanged to the holdout set. The resulting holdout \(\tau\)-nRMSE is reported in Table~\ref{tab:r11_segmentation_losses}. This is a scalar-readout choice, not probabilistic calibration; likelihood, coverage, and posterior concentration are evaluated separately below.

\subsection{Posterior Geometry Diagnostics}

Once the scalar readout has been fixed, the event-time posterior enables evaluation beyond point prediction because similar nRMSE values can correspond to substantially different posterior geometry. We therefore evaluate Width80 (central 80\% interval width), Mass \(\pm150\) ms (probability assigned within 150 ms of observed RT), and mode--mean disagreement (absolute distance between posterior mode and mean), measuring temporal concentration, target alignment, and agreement with the expectation used for scalar prediction. Width80 and mode--mean disagreement are summarized by their within-seed holdout medians, and target-aligned mass by its mean.

\begin{figure}[!t]
    \centering
    \includegraphics[width=0.88\linewidth]{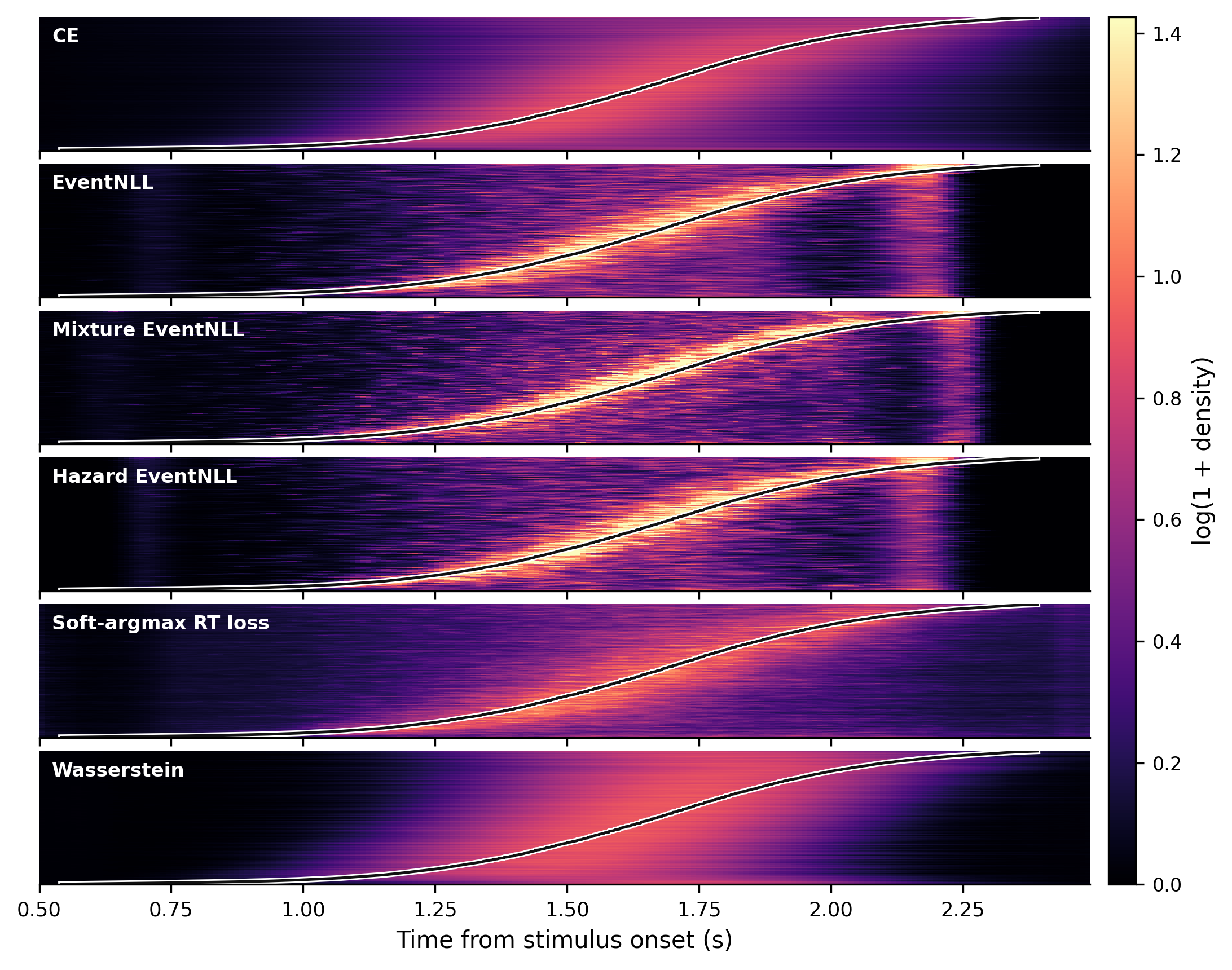}
    \caption{Trial-sorted event-time posterior maps on the holdout split for matched segmentation losses. Rows are quantile bins of support-filtered holdout trials sorted by observed RT; the x-axis is time from stimulus onset; color shows log-transformed posterior density averaged within each bin. The overlaid curve marks the mean observed RT in each bin.}
    \label{fig:posterior_raster_main}
\end{figure}

In Figure~\ref{fig:posterior_raster_main}, CE and Wasserstein place broad probability mass across the response window, whereas EventNLL-family objectives produce sharper bands that more closely track the observed RT curve.

\begin{figure}[!t]
    \centering
    \includegraphics[width=\linewidth]{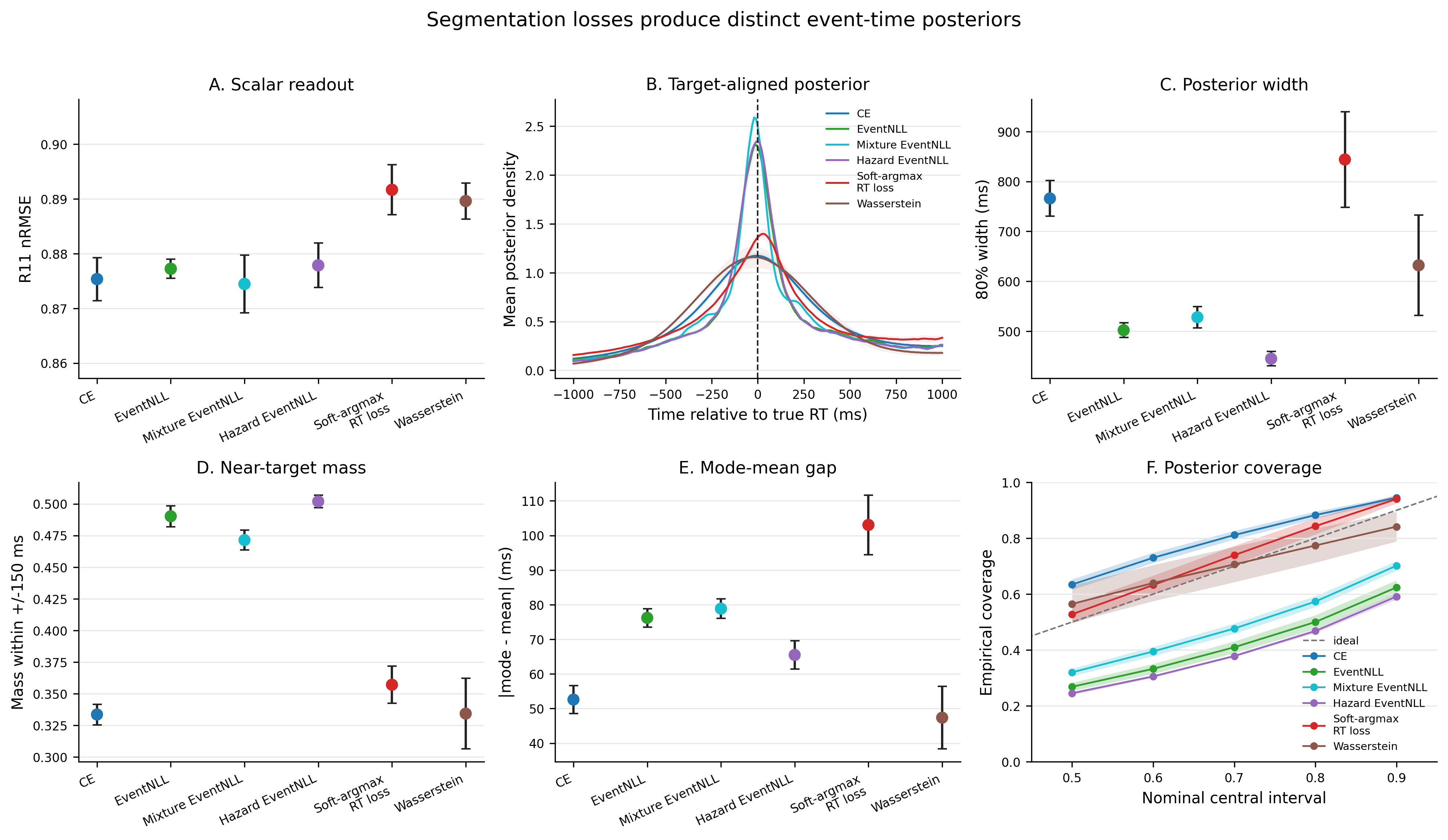}
    \caption{Posterior geometry differs across event-time losses even when scalar nRMSE is similar. Panel A reports holdout scalar readout after selecting the objective-specific readout temperature on R9--R10 to minimize nRMSE. Panel B aligns predicted posteriors to the observed RT. Panels C--E summarize posterior width, near-target mass, and mode-mean disagreement. Panel F compares empirical coverage of central posterior intervals; the diagonal is ideal interval coverage.}
    \label{fig:posterior_geometry_main}
\end{figure}

Figure~\ref{fig:posterior_geometry_main} summarizes these posterior-geometry differences, while Table~\ref{tab:posterior_geometry_metrics} reports scalar accuracy, shared-kernel RT NLL, width, mass, and coverage. Shared-kernel RT NLL provides a common likelihood-based score across objectives with different training scales; its parameters are reported in Table~\ref{tab:reproducibility_details}. Coverage80 is the fraction of observed RTs inside the central 80\% posterior interval, and Coverage MAE averages the absolute nominal--empirical coverage difference over central interval levels of 50\%, 60\%, 70\%, 80\%, and 90\%. Together, these metrics separate scalar accuracy, distributional scoring, temporal concentration, target alignment, and interval behavior.

\begin{table}[!htbp]
\centering
\caption{Holdout posterior-geometry metrics across event-time segmentation losses.}
\label{tab:posterior_geometry_metrics}
\scriptsize
\setlength{\tabcolsep}{3pt}
\renewcommand{\arraystretch}{1.12}
\resizebox{\linewidth}{!}{%
\begin{tabular}{@{} l c c c c c c @{}}
\toprule
\textbf{Objective} & \textbf{nRMSE $\downarrow$} & \textbf{Shared-kernel RT NLL $\downarrow$} & \textbf{Width80 ms $\downarrow$} & \textbf{Mass $\pm$150 ms $\uparrow$} & \textbf{Coverage80} & \textbf{Coverage MAE $\downarrow$} \\
\midrule
CE & 0.875 $\pm$ 0.004 & 0.08 $\pm$ 0.01 & 766 $\pm$ 36 & 0.334 $\pm$ 0.008 & 0.883 $\pm$ 0.013 & 0.101 $\pm$ 0.014 \\
EventNLL & 0.877 $\pm$ 0.002 & -0.06 $\pm$ 0.01 & 502 $\pm$ 15 & 0.490 $\pm$ 0.008 & 0.500 $\pm$ 0.025 & 0.273 $\pm$ 0.021 \\
Mixture EventNLL & \textbf{0.874 $\pm$ 0.005} & \textbf{-0.08 $\pm$ 0.01} & 528 $\pm$ 22 & 0.471 $\pm$ 0.008 & 0.573 $\pm$ 0.020 & 0.207 $\pm$ 0.017 \\
Hazard EventNLL & 0.878 $\pm$ 0.004 & -0.06 $\pm$ 0.01 & \textbf{445 $\pm$ 14} & \textbf{0.502 $\pm$ 0.005} & 0.468 $\pm$ 0.007 & 0.302 $\pm$ 0.006 \\
Soft-argmax RT loss & 0.892 $\pm$ 0.005 & 0.11 $\pm$ 0.03 & 844 $\pm$ 96 & 0.357 $\pm$ 0.015 & 0.843 $\pm$ 0.029 & \textbf{0.040 $\pm$ 0.024} \\
Wasserstein & 0.890 $\pm$ 0.003 & 0.20 $\pm$ 0.04 & 632 $\pm$ 101 & 0.334 $\pm$ 0.028 & \textbf{0.774 $\pm$ 0.061} & 0.059 $\pm$ 0.021 \\
\bottomrule
\end{tabular}%
}
\end{table}

These diagnostics show that losses with similar scalar error can induce different posterior semantics. CE and mixture EventNLL provide the strongest scalar readouts. EventNLL-family objectives are sharper and more target-concentrated, but their latent-posterior intervals have low empirical coverage. The soft-argmax RT-loss control combines the lowest Coverage MAE with the broadest posterior, but it is weaker as a scalar predictor and lacks distributional event-time supervision. Wasserstein is closest to nominal 80\% coverage, but it has weaker scalar accuracy and lower near-target mass. Thus, posterior geometry reveals trade-offs among scalar accuracy, target concentration, distributional scoring, and interval behavior that scalar nRMSE alone would hide.

Accordingly, CE and mixture EventNLL are strongest for scalar RT prediction in this study, whereas EventNLL-family objectives provide sharper, more target-concentrated latent posteriors. Wasserstein and the soft-argmax control yield latent-posterior intervals closer to nominal coverage, but these intervals are not calibrated predictive RT uncertainty; the following analysis evaluates that distinction through the observation-noise model.

\paragraph{Latent Event-Time Posterior and Predictive RT Calibration.}
The coverage columns in Table~\ref{tab:posterior_geometry_metrics} report how often observed RT falls inside central intervals of the latent event-time posterior, \(p(t_{\mathrm{event}}\mid X)\). For EventNLL-family models, the predictive distribution over observed RT is instead obtained by combining this posterior with an observation kernel:
\[
p(y_{\mathrm{RT}}\mid X)=\sum_t p(t_{\mathrm{event}}=t\mid X)\,K(y_{\mathrm{RT}}\mid t).
\]
This separates uncertainty over the latent response-relevant event time from predictive uncertainty over observed RT. Appendix Table~\ref{tab:observation_noise_calibration} evaluates predictive calibration while keeping the trained EEG model, event-time posterior, and posterior-mean prediction fixed. A single multiplicative scale for the observation kernel is selected on R9--R10 by Coverage MAE over central intervals from 50\% to 90\% and applied unchanged to holdout. This reduces predictive RT Coverage MAE from 0.040--0.059 to 0.005--0.008, with holdout Coverage80 near nominal coverage (0.790--0.795).

The latent posterior therefore describes event-time concentration, whereas the observation model supplies calibrated predictive RT intervals without altering point predictions. We next test whether posterior predictions move in a crop-relative way when the temporal frame is shifted.

\subsection{Shifted-Crop Shortcut-vs-Localization Diagnostic}
\label{sec:shifted_crop_diagnostic}

Posterior geometry characterizes probability mass within the standard stimulus-locked window, but it does not test whether predictions behave as crop-relative timing localizers when the temporal frame changes. A fixed-window model may instead rely on trial difficulty, subject-level response tendency, or stimulus-locked timing structure.

We therefore use shifted-crop inference as a shortcut-vs-localization diagnostic on the holdout split. For each trial, the trained model is evaluated on 2~s crops from the same 5~s EEG segment, starting at $s\in\{0.2,0.3,\dots,0.8\}$~s after stimulus onset. The canonical window starts at $s=0.5$~s, and the target for crop $s$ is $\mathrm{RT}-s$. A crop-relative localizer should adjust its prediction when the crop start changes, whereas a model tied to stimulus-locked scalar timing should be less sensitive.

Table~\ref{tab:shift_jitter_summary} summarizes this diagnostic and the matched shift-jitter intervention. Shifted relative nRMSE measures crop-relative accuracy over pooled shifted-crop examples for which the response remains inside the evaluated 2~s window. Shift sensitivity compares the magnitude of the prediction change from the canonical crop with the imposed crop shift; 1 indicates ideal crop-relative responsiveness and 0 indicates crop-invariant prediction. Direction agreement is the fraction of examples for which prediction changes opposite to the crop shift. Sensitivity and direction are computed on the common trial subset whose responses remain inside every evaluated crop, ensuring comparisons over the same trials and crop starts.

\begin{table}[!t]
\centering
\caption{Shifted-crop diagnostic and shift-jitter intervention on the 5~s holdout dataset. Values are seed means; full mean $\pm$ standard deviation values are reported in Appendix Table~\ref{tab:shift_jitter_summary_full}. Arrows indicate the preferred direction.}
\label{tab:shift_jitter_summary}
\footnotesize
\setlength{\tabcolsep}{3.5pt}
\renewcommand{\arraystretch}{1.12}
\begin{tabular}{@{} l c c c c c c c c @{}}
\toprule
\textbf{Objective} & \multicolumn{2}{c}{\textbf{Holdout \(\tau\)-nRMSE} $\downarrow$} & \multicolumn{2}{c}{\textbf{Shifted rel. nRMSE} $\downarrow$} & \multicolumn{2}{c}{\textbf{Sensitivity} $\uparrow$} & \multicolumn{2}{c}{\textbf{Direction} $\uparrow$} \\
\cmidrule(lr){2-3}\cmidrule(lr){4-5}\cmidrule(lr){6-7}\cmidrule(l){8-9}
 & \textbf{Fixed} & \textbf{Jitter} & \textbf{Fixed} & \textbf{Jitter} & \textbf{Fixed} & \textbf{Jitter} & \textbf{Fixed} & \textbf{Jitter} \\
\midrule
CE & 0.8753 & 0.8749 & 0.8679 & \textbf{0.8569} & 0.5810 & 0.5831 & 0.7780 & 0.7924 \\
Mixture EventNLL & \textbf{0.8745} & \textbf{0.8734} & \textbf{0.8663} & 0.8576 & 0.6021 & 0.6249 & 0.7730 & 0.7925 \\
EventNLL & 0.8772 & 0.8771 & 0.8685 & 0.8593 & 0.5842 & 0.6174 & 0.7739 & 0.7937 \\
Hazard EventNLL & 0.8778 & 0.8806 & 0.8692 & 0.8609 & 0.5618 & 0.5891 & 0.7694 & 0.7925 \\
Soft-argmax RT loss & 0.8917 & 0.8836 & 0.8857 & 0.8613 & 0.5381 & 0.5775 & 0.7592 & 0.7951 \\
Wasserstein & 0.8896 & 0.8922 & 0.8932 & 0.8774 & \textbf{0.6684} & \textbf{0.6852} & \textbf{0.7830} & \textbf{0.8026} \\
\bottomrule
\end{tabular}
\end{table}

We then trained matched shift-jitter variants by sampling the 2~s training window from the same crop-start range and expressing the target relative to the sampled window. For CE, mixture EventNLL, and Gaussian EventNLL, the intervention preserves standard fixed-window holdout accuracy. Shifted relative nRMSE improves for every evaluated objective. Mean direction agreement and sensitivity also increase for every objective, showing that predictions move more consistently in the expected localizer direction. Sensitivity remains below 1, quantifying the residual equivariance gap.

Across objectives, Wasserstein shows the strongest crop-relative sensitivity and direction agreement but weaker scalar and shifted-crop accuracy, illustrating a trade-off between crop-relative movement and point prediction.

\FloatBarrier

\section{Discussion}

\subsection{Latency Targets as Output Representations}

For latency-labeled EEG, the supervised target representation is itself a modeling choice rather than a neutral readout convention. The label is a behavioral latency within a post-stimulus interval, and models benefit when that temporal structure is explicit in the output space and in the loss. CE and the EventNLL-family objectives form the strongest scalar-readout group, whereas the soft-argmax RT-loss control is weaker despite using the same posterior-mean readout. This supports the interpretation that the gain comes from distributional event-time supervision, not merely from replacing a scalar head with a differentiable expectation.

Objective choice therefore shapes not only scalar accuracy but also the structure of the learned posterior. CE-style soft targets provide strong scalar readouts and broad posterior support. EventNLL reaches a similar scalar-error regime through a latent-variable likelihood, marginalizing an unobserved response-relevant event time under an observation model for RT. Wasserstein behaves differently: it is more localizer-like by shifted-crop sensitivity and direction, but weaker as a point predictor. Thus, the choice of objective controls a trade-off among scalar accuracy, posterior concentration, interval behavior, and crop-relative movement.

\subsection{Posterior Diagnostics Beyond nRMSE}

The event-time posterior contains more information than its expectation. Posterior width, target-aligned mass, mode-mean disagreement, shared-kernel RT NLL, and interval coverage expose temporal output behavior that scalar nRMSE discards. This is important because similar point errors can correspond to different posterior semantics: broad conservative distributions, sharper target-concentrated distributions, or better interval coverage with weaker scalar prediction.

Temperature tuning serves a specific role in this protocol: it standardizes posterior-mean predictions for scalar nRMSE comparison. Distributional diagnostics remain separate: likelihood, coverage, posterior concentration, and temporal-perturbation behavior answer different questions about the learned event-time posterior. The observation-noise calibration analysis further separates three goals that are easy to conflate: scalar RT accuracy, latent event-time concentration, and calibrated predictive uncertainty over observed behavioral RT.

\subsection{Temporal Perturbation and Shortcut Behavior}

The shifted-crop diagnostic asks whether posterior predictions behave as temporal localizers when the crop frame changes. If models rely on fixed-window, stimulus-locked shortcuts, crop shifts should expose this by separating scalar accuracy from crop-relative movement. Shift-jitter training addresses this behavior by varying the crop start during training and expressing the target relative to the sampled crop. Across every evaluated objective, shift-jitter improves shifted-crop accuracy and expected-direction movement while preserving canonical-window accuracy for the primary event-time models. Mean sensitivity and direction increase across all objectives, demonstrating that crop-relative behavior can be strengthened through target-aligned temporal augmentation. Sensitivity below 1 shows that localization remains partial and provides a quantitative target for fully equivariant models.

\subsection{Scope, Limitations, and Generality}

The external architecture baselines place the formulation result in context: under the same preprocessing and normalization protocol, compact task-specific temporal models outperform the evaluated standardized EEG backbones and foundation-style architectures trained from scratch. Across matched dense temporal backbones, distributionally supervised objectives also outperform RT-only posterior-mean training. Thus, within this benchmark, these complementary controls identify formulation and objective design as a consistent source of predictive improvement beyond architecture capacity alone.

The empirical scope is the HBN-EEG contrast change detection task under a release-separated protocol. Broader generality remains to be tested across tasks, acquisition settings, montages, and latency targets. The shifted-crop analysis provides response-timing localization evidence and also shows that full crop-relative localization remains an open target for objectives, architectures, or augmentations designed for temporal equivariance.

\section{Conclusion}

Reaction time is usually evaluated as a scalar, but it also specifies when a behavioral response occurs. We therefore treated RT as weak event-time supervision and learned a posterior over response-relevant timing from single-trial EEG. Under the subject-disjoint, release-separated HBN-EEG protocol, distributional event-time supervision consistently improved held-out RT prediction over direct scalar regression and an RT-only posterior-mean control. Five-seed comparisons and replication across four dense temporal backbones show that this advantage is attributable to the supervised output formulation rather than to model scale, expectation readout, or a particular architecture.

The posterior representation also changed what could be evaluated. CE and the EventNLL-family objectives achieved the strongest scalar readouts but induced different posterior geometries; observation-noise calibration separated latent event-time concentration from predictive uncertainty over observed RT; and shifted-crop analyses revealed partial crop-relative localization that scalar nRMSE alone cannot diagnose. Shift-jitter strengthened shifted-crop robustness and expected-direction movement, while the remaining sensitivity gap defines a clear target for temporally equivariant modeling.

The broader implication is that latency labels need not be reduced to scalar endpoints. They can serve as structured supervision that preserves temporal meaning while remaining compatible with standard behavioral metrics. Extending this view to other tasks and latency-defined EEG phenomena could support models that predict behavior accurately, expose the temporal organization of their outputs, and make their reliance on temporal evidence directly testable.

\appendix

\section{Additional Readout, Calibration, and Shifted-Crop Details}

This appendix records three protocol details that are secondary to the main results but important for reproducing the posterior diagnostics. First, post-hoc observation-noise calibration separates latent event-time posterior coverage from behavioral-RT predictive coverage. Second, the posterior readout temperature is selected only on the R9--R10 development split and then applied unchanged to the holdout split. Third, the shifted-crop table in the main text reports seed means for readability; the full seed-variability summary is given here.

\paragraph{Observation-noise calibration.}
Table~\ref{tab:observation_noise_calibration} evaluates a post-hoc RT observation-noise calibration for EventNLL-family models. The trained EEG model, event-time posterior, and posterior-mean scalar readout are fixed. The latent posterior rows evaluate central intervals of \(p(t_{\mathrm{event}}\mid X)\) directly, whereas the predictive RT rows convolve the same posterior with either the original observation kernel or a calibrated kernel scale selected on R9--R10 by Coverage MAE across central intervals 50--90\%. This calibration changes only the RT observation-noise layer used for probabilistic prediction; it does not change trained weights, posterior-mean RT predictions, or \(\tau\)-nRMSE.

\begin{figure}[!t]
    \centering
    \includegraphics[width=0.9\linewidth]{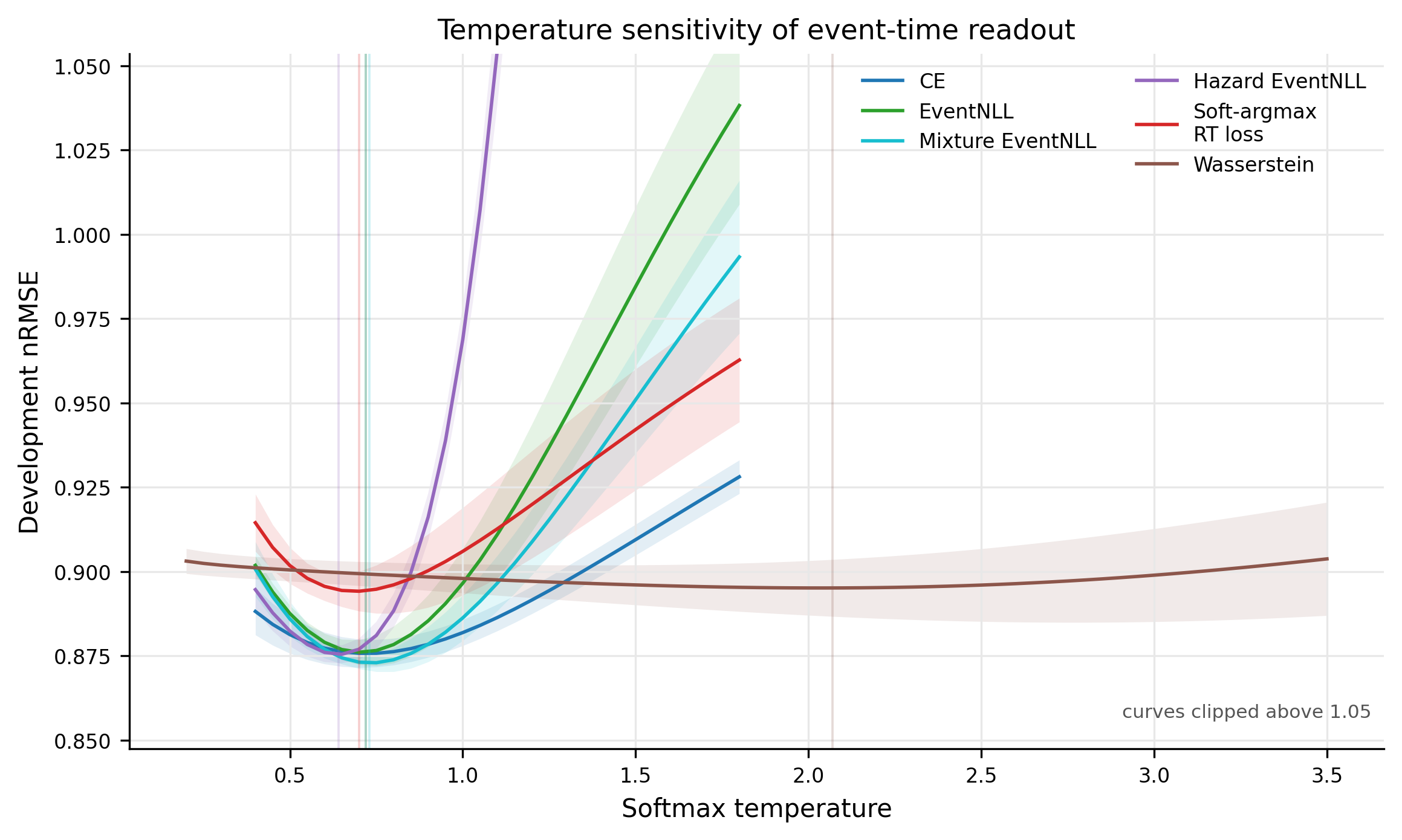}
    \caption{Readout-temperature sensitivity on the R9--R10 development split. Curves show posterior-mean nRMSE as a function of softmax temperature, and vertical lines mark selected temperatures. The selected temperature is then applied unchanged to holdout predictions and posterior diagnostics.}
    \label{fig:appendix_temperature_sensitivity}
\end{figure}

\paragraph{Readout-temperature selection.}
Figure~\ref{fig:appendix_temperature_sensitivity} shows the development-split nRMSE curves used to select the posterior-readout temperature for each event-time objective. The minima differ across losses, confirming that objectives induce different posterior confidence scales. We therefore treat temperature selection as part of the scalar readout protocol before comparing posterior-mean predictions and posterior geometry on the holdout split.

\begin{table}[!t]
\centering
\caption{Post-hoc RT observation-noise calibration for EventNLL-family posteriors. Coverage MAE is averaged over central interval levels 50\%, 60\%, 70\%, 80\%, and 90\%.}
\label{tab:observation_noise_calibration}
\scriptsize
\setlength{\tabcolsep}{4pt}
\renewcommand{\arraystretch}{1.12}
\resizebox{\linewidth}{!}{%
\begin{tabular}{@{} l c c c c c @{}}
\toprule
\textbf{Distribution evaluated} & \textbf{Scale \(c\)} & \textbf{Coverage MAE $\downarrow$} & \textbf{Coverage80} & \textbf{Width80 ms} & \textbf{Predictive NLL $\downarrow$} \\
\midrule
\multicolumn{6}{@{}l}{\textit{EventNLL}} \\
\addlinespace[0.15em]
Latent event-time posterior & -- & 0.273 $\pm$ 0.021 & 0.500 $\pm$ 0.025 & 502 $\pm$ 15 & -- \\
Predictive RT, original \(K\) & 1.00 & 0.059 $\pm$ 0.004 & 0.850 $\pm$ 0.004 & 699 $\pm$ 13 & -0.044 $\pm$ 0.010 \\
Predictive RT, calibrated \(K_c\) & 0.70 $\pm$ 0.00 & 0.006 $\pm$ 0.003 & 0.793 $\pm$ 0.004 & 632 $\pm$ 14 & -0.054 $\pm$ 0.016 \\
\midrule
\multicolumn{6}{@{}l}{\textit{Mixture EventNLL}} \\
\addlinespace[0.15em]
Latent event-time posterior & -- & 0.207 $\pm$ 0.017 & 0.573 $\pm$ 0.020 & 528 $\pm$ 22 & -- \\
Predictive RT, original \(K\) & 1.00 & 0.040 $\pm$ 0.010 & 0.835 $\pm$ 0.009 & 671 $\pm$ 17 & -0.098 $\pm$ 0.012 \\
Predictive RT, calibrated \(K_c\) & 0.73 $\pm$ 0.04 & 0.005 $\pm$ 0.004 & 0.795 $\pm$ 0.003 & 622 $\pm$ 12 & -0.111 $\pm$ 0.013 \\
\midrule
\multicolumn{6}{@{}l}{\textit{Hazard EventNLL}} \\
\addlinespace[0.15em]
Latent event-time posterior & -- & 0.303 $\pm$ 0.006 & 0.467 $\pm$ 0.007 & 440 $\pm$ 17 & -- \\
Predictive RT, original \(K\) & 1.00 & 0.044 $\pm$ 0.006 & 0.834 $\pm$ 0.006 & 639 $\pm$ 15 & -0.044 $\pm$ 0.006 \\
Predictive RT, calibrated \(K_c\) & 0.78 $\pm$ 0.04 & 0.008 $\pm$ 0.002 & 0.790 $\pm$ 0.005 & 587 $\pm$ 17 & -0.059 $\pm$ 0.005 \\
\bottomrule
\end{tabular}%
}
\end{table}

\FloatBarrier

\paragraph{Shifted-crop seed variability.}
Table~\ref{tab:shift_jitter_summary_full} expands the main shifted-crop diagnostic table by reporting mean $\pm$ standard deviation across seeds. The seed-level results reproduce the main-text pattern: shift-jitter lowers mean shifted-crop relative nRMSE and increases both mean sensitivity and direction for every evaluated objective. These consistent changes show that shift-jitter moves predictions toward crop-relative temporal localization across loss families. Absolute sensitivity remains below ideal equivariance, quantifying the localization gap that remains to be addressed.

\begin{table}[!t]
\centering
\caption{Full shifted-crop diagnostic and shift-jitter intervention summary with seed variability. Values are mean $\pm$ standard deviation across seeds; the corresponding main-text table reports means only for readability.}
\label{tab:shift_jitter_summary_full}
\scriptsize
\setlength{\tabcolsep}{2.2pt}
\renewcommand{\arraystretch}{1.12}
\resizebox{\linewidth}{!}{%
\begin{tabular}{@{} l c c c c c c c c @{}}
\toprule
\textbf{Objective} & \multicolumn{2}{c}{\textbf{Holdout \(\tau\)-nRMSE} $\downarrow$} & \multicolumn{2}{c}{\textbf{Shifted rel. nRMSE} $\downarrow$} & \multicolumn{2}{c}{\textbf{Sensitivity} $\uparrow$} & \multicolumn{2}{c}{\textbf{Direction} $\uparrow$} \\
\cmidrule(lr){2-3}\cmidrule(lr){4-5}\cmidrule(lr){6-7}\cmidrule(l){8-9}
 & \textbf{Fixed} & \textbf{Jitter} & \textbf{Fixed} & \textbf{Jitter} & \textbf{Fixed} & \textbf{Jitter} & \textbf{Fixed} & \textbf{Jitter} \\
\midrule
CE & 0.8753 $\pm$ 0.0039 & 0.8749 $\pm$ 0.0060 & 0.8679 $\pm$ 0.0024 & 0.8569 $\pm$ 0.0060 & 0.5810 $\pm$ 0.0351 & 0.5831 $\pm$ 0.0406 & 0.7780 $\pm$ 0.0041 & 0.7924 $\pm$ 0.0056 \\
Mixture EventNLL & 0.8745 $\pm$ 0.0053 & 0.8734 $\pm$ 0.0033 & 0.8663 $\pm$ 0.0037 & 0.8576 $\pm$ 0.0041 & 0.6021 $\pm$ 0.0484 & 0.6249 $\pm$ 0.0313 & 0.7730 $\pm$ 0.0062 & 0.7925 $\pm$ 0.0076 \\
EventNLL & 0.8772 $\pm$ 0.0018 & 0.8771 $\pm$ 0.0040 & 0.8685 $\pm$ 0.0023 & 0.8593 $\pm$ 0.0028 & 0.5842 $\pm$ 0.0287 & 0.6174 $\pm$ 0.0222 & 0.7739 $\pm$ 0.0059 & 0.7937 $\pm$ 0.0068 \\
Hazard EventNLL & 0.8778 $\pm$ 0.0041 & 0.8806 $\pm$ 0.0034 & 0.8692 $\pm$ 0.0028 & 0.8609 $\pm$ 0.0045 & 0.5618 $\pm$ 0.0376 & 0.5891 $\pm$ 0.0259 & 0.7694 $\pm$ 0.0090 & 0.7925 $\pm$ 0.0074 \\
Soft-argmax RT loss & 0.8917 $\pm$ 0.0046 & 0.8836 $\pm$ 0.0035 & 0.8857 $\pm$ 0.0069 & 0.8613 $\pm$ 0.0032 & 0.5381 $\pm$ 0.0586 & 0.5775 $\pm$ 0.0362 & 0.7592 $\pm$ 0.0183 & 0.7951 $\pm$ 0.0052 \\
Wasserstein & 0.8896 $\pm$ 0.0033 & 0.8922 $\pm$ 0.0066 & 0.8932 $\pm$ 0.0033 & 0.8774 $\pm$ 0.0099 & 0.6684 $\pm$ 0.0711 & 0.6852 $\pm$ 0.0416 & 0.7830 $\pm$ 0.0197 & 0.8026 $\pm$ 0.0118 \\
\bottomrule
\end{tabular}%
}
\end{table}

\FloatBarrier

\section{Architecture-Control Details}
\label{app:architecture_controls}

\paragraph{Architecture specifications.}
The architecture-control hyperparameters and parameter counts are summarized
in Table~\ref{tab:reproducibility_details}.

\paragraph{Diagnostic robustness.}
Table~\ref{tab:architecture_control_diagnostics} complements the fixed-window
accuracy comparison in Table~\ref{tab:architecture_robustness} by testing
whether shifted-crop behavior and objective-dependent posterior geometry also
persist across backbones. The posterior columns retain a common distributional
score, target-aligned mass, and interval coverage error.

\begin{table}[!htbp]
\centering
\caption{Shifted-crop and posterior-geometry diagnostics across dense temporal backbones. Shifted-crop metrics follow the evaluation convention in Section~\ref{sec:shifted_crop_diagnostic}; posterior metrics are evaluated on the fixed holdout window after development-set readout-temperature selection. Values are mean $\pm$ standard deviation across five seeds.}
\label{tab:architecture_control_diagnostics}
\scriptsize
\setlength{\tabcolsep}{2.2pt}
\renewcommand{\arraystretch}{1.12}
\resizebox{\linewidth}{!}{%
\begin{tabular}{@{} l c c c c c c @{}}
\toprule
\textbf{Objective}
& \multicolumn{3}{c}{\textbf{Shifted-crop diagnostics}}
& \multicolumn{3}{c}{\textbf{Posterior geometry}} \\
\cmidrule(lr){2-4}\cmidrule(l){5-7}
& \textbf{Rel. nRMSE $\downarrow$}
& \textbf{Sensitivity $\uparrow$}
& \textbf{Direction $\uparrow$}
& \textbf{Shared-kernel RT NLL $\downarrow$}
& \textbf{Mass $\pm150$ ms $\uparrow$}
& \textbf{Coverage MAE $\downarrow$} \\
\midrule
\multicolumn{7}{@{}l}{\textit{ETS-U-Net}} \\
\addlinespace[0.15em]
RT-only soft-argmax & 0.886 $\pm$ 0.007 & 0.538 $\pm$ 0.059 & 0.759 $\pm$ 0.018 & 0.1070 $\pm$ 0.0303 & 0.357 $\pm$ 0.015 & 0.040 $\pm$ 0.024 \\
CE & 0.868 $\pm$ 0.002 & 0.581 $\pm$ 0.035 & 0.778 $\pm$ 0.004 & 0.0770 $\pm$ 0.0144 & 0.334 $\pm$ 0.008 & 0.101 $\pm$ 0.014 \\
Mixture EventNLL & 0.866 $\pm$ 0.004 & 0.602 $\pm$ 0.048 & 0.773 $\pm$ 0.006 & -0.0820 $\pm$ 0.0122 & 0.471 $\pm$ 0.008 & 0.207 $\pm$ 0.017 \\
\midrule
\multicolumn{7}{@{}l}{\textit{ETS-TCN}} \\
\addlinespace[0.15em]
RT-only soft-argmax & 0.881 $\pm$ 0.006 & 0.519 $\pm$ 0.042 & 0.758 $\pm$ 0.012 & 0.0268 $\pm$ 0.0114 & 0.410 $\pm$ 0.012 & 0.080 $\pm$ 0.031 \\
CE & 0.870 $\pm$ 0.008 & 0.551 $\pm$ 0.025 & 0.774 $\pm$ 0.006 & 0.0710 $\pm$ 0.0243 & 0.336 $\pm$ 0.013 & 0.099 $\pm$ 0.016 \\
Mixture EventNLL & 0.867 $\pm$ 0.005 & 0.544 $\pm$ 0.036 & 0.761 $\pm$ 0.010 & -0.0835 $\pm$ 0.0039 & 0.471 $\pm$ 0.007 & 0.187 $\pm$ 0.017 \\
\midrule
\multicolumn{7}{@{}l}{\textit{ETS-InceptionPyramid}} \\
\addlinespace[0.15em]
RT-only soft-argmax & 0.882 $\pm$ 0.005 & 0.506 $\pm$ 0.035 & 0.758 $\pm$ 0.007 & 0.1634 $\pm$ 0.0195 & 0.318 $\pm$ 0.007 & 0.103 $\pm$ 0.020 \\
CE & 0.863 $\pm$ 0.004 & 0.533 $\pm$ 0.032 & 0.781 $\pm$ 0.012 & 0.0605 $\pm$ 0.0173 & 0.340 $\pm$ 0.012 & 0.097 $\pm$ 0.017 \\
Mixture EventNLL & 0.866 $\pm$ 0.002 & 0.589 $\pm$ 0.037 & 0.774 $\pm$ 0.005 & -0.0925 $\pm$ 0.0049 & 0.473 $\pm$ 0.008 & 0.185 $\pm$ 0.019 \\
\midrule
\multicolumn{7}{@{}l}{\textit{ETS-AttnSeg}} \\
\addlinespace[0.15em]
RT-only soft-argmax & 0.899 $\pm$ 0.033 & 0.479 $\pm$ 0.156 & 0.749 $\pm$ 0.028 & 0.2977 $\pm$ 0.1669 & 0.268 $\pm$ 0.048 & 0.156 $\pm$ 0.046 \\
CE & 0.868 $\pm$ 0.009 & 0.566 $\pm$ 0.057 & 0.771 $\pm$ 0.005 & 0.0771 $\pm$ 0.0276 & 0.334 $\pm$ 0.016 & 0.103 $\pm$ 0.019 \\
Mixture EventNLL & 0.867 $\pm$ 0.008 & 0.639 $\pm$ 0.034 & 0.767 $\pm$ 0.004 & -0.0937 $\pm$ 0.0083 & 0.477 $\pm$ 0.007 & 0.173 $\pm$ 0.023 \\
\bottomrule
\end{tabular}%
}
\end{table}

Across every backbone, CE and mixture EventNLL consistently achieve lower mean
shifted-crop error than RT-only posterior-mean training and produce more
crop-responsive predictions, as reflected in higher sensitivity and direction
scores. These improvements replicate across all four temporal architectures.
Absolute sensitivity remains below 1 and therefore quantifies partial rather
than fully crop-relative localization. Mixture EventNLL shows an equally
consistent posterior-geometry pattern: within every backbone, it yields the
lowest shared-kernel RT NLL and the greatest target-aligned mass, together with
higher coverage error. The replication of both patterns across four distinct
dense temporal backbones establishes the architecture robustness of the
supervision effect. These experiments serve as controlled robustness checks
rather than a ranking of general EEG architectures.

\FloatBarrier

\section{Reproducibility Details}
\label{app:reproducibility_details}

YAML configuration files at \url{https://github.com/sneddy/neurosned/tree/main/benchmarks/configs} specify all reported experiments. Training summaries, prediction files, selected temperatures, and model summaries are written to \href{https://github.com/sneddy/neurosned/tree/main/benchmarks/experiments}{\texttt{benchmarks/experiments/}}. Table~\ref{tab:reproducibility_details} summarizes the shared settings needed to reproduce the controlled comparisons: data splits, input representation, optimization, augmentation, model selection, readout-temperature tuning, and the main architecture hyperparameters. Settings not listed in the table are objective-specific loss choices described in the main text.

\begin{table}[!htbp]
\centering
\caption{Reproducibility summary for the main controlled comparisons.}
\label{tab:reproducibility_details}
\scriptsize
\setlength{\tabcolsep}{4pt}
\renewcommand{\arraystretch}{1.12}
\begin{tabularx}{\linewidth}{@{} p{0.17\linewidth} p{0.25\linewidth} X @{}}
\toprule
\textbf{Group} & \textbf{Setting} & \textbf{Value} \\
\midrule
\multicolumn{3}{@{}l}{\textit{Data and splits}} \\
\addlinespace[0.15em]
Input window & EEG segment & 0.5--2.5~s after stimulus onset; 2~s, 200 samples, 100~Hz, 128 channels. \\
Target support & Main analyzed set & \(0.5 \leq \mathrm{RT} \leq 2.5\)~s. \\
Splits & Release-separated protocol & R1--R8 train, R9--R10 development, R11 final holdout. \\
Seeds & Repeated runs & 2025--2029. \\
\midrule
\multicolumn{3}{@{}l}{\textit{Optimization and selection}} \\
\addlinespace[0.15em]
Optimizer & Neural models & Adam, learning rate \(10^{-3}\), weight decay 0. \\
Training loop & Epochs and batch size & Maximum 100 epochs; train batch size 128. \\
Evaluation loader & Batch size & Development and holdout batch size 256. \\
Checkpoint selection & Model selection & Best development nRMSE; early stopping patience 20. \\
Holdout use & Final evaluation & R11 is evaluated after checkpoint and readout choices are fixed. \\
Confidence intervals & Holdout summaries & Subject bootstrap over R11 subjects; 1000 samples for per-run intervals and 5000 samples for the scalar effect-size analysis. \\
\midrule
\multicolumn{3}{@{}l}{\textit{Training augmentations}} \\
\addlinespace[0.15em]
Channel dropout & Shared training augmentation & Applied with probability 0.25; dropped-channel ratio sampled uniformly from 0 to 0.3. \\
Temporal cutout & Shared training augmentation & Applied with probability 0.25; contiguous zeroed segment length 10--50 samples. \\
Gaussian noise & Shared training augmentation & Applied with probability 0.30; noise standard deviation \(0.01 + 0.01|\mathcal{N}(0,1)|\). \\
Shift-jitter & Intervention experiments only & Training crop start sampled from 0.2--0.8~s, crop length 2~s, target expressed relative to crop start. \\
\bottomrule
\end{tabularx}
\end{table}

\begin{table}[!htbp]
\centering
\begingroup
\renewcommand{\theHtable}{\arabic{table}-continued}
\renewcommand{\addcontentsline}[3]{}
\addtocounter{table}{-1}
\caption{Reproducibility summary for the main controlled comparisons (continued).}
\endgroup
\scriptsize
\setlength{\tabcolsep}{4pt}
\renewcommand{\arraystretch}{1.12}
\begin{tabularx}{\linewidth}{@{} p{0.17\linewidth} p{0.25\linewidth} X @{}}
\toprule
\textbf{Group} & \textbf{Setting} & \textbf{Value} \\
\midrule
\multicolumn{3}{@{}l}{\textit{Event-time readout}} \\
\addlinespace[0.15em]
Event grid & Temporal support & 200 bins over 2~s; 10~ms per bin. \\
Soft target & Gaussian event-time label & \(\sigma=0.15\)~s. \\
Likelihood-based objectives & Observation kernels & \(\sigma_y=0.15\)~s: \(\sigma_{\mathrm{narrow}}=0.10\)~s, \(\sigma_{\mathrm{wide}}=0.35\)~s, \(\alpha=0.10\). \\
Posterior diagnostics & Shared-kernel RT NLL & \(\sigma_{\mathrm{score}}=0.12\)~s across objectives. \\
Posterior readout & Scalar prediction & Posterior mean / soft-argmax. \\
Base temperature & Training/evaluation default & \(\tau=0.65\) before post-hoc readout-temperature selection. \\
Temperature selection & Readout tuning & Grid search on R9--R10 minimizing posterior-mean nRMSE; selected \(\tau\) applied unchanged to R11. \\
Temperature grid & Objective-specific grid & \(0.2\)--\(3.5\) with step 0.05. \\
\midrule
\multicolumn{3}{@{}l}{\textit{Model architectures}} \\
\addlinespace[0.15em]
ETS-U-Net & Primary event-time backbone & \(c_0=96\), widen 2, depth per stage 5, kernel size 15, dropout 0.2, one output channel; 3.10M parameters. \\
ETS-TCN & Dilated-convolution control & \(c_0=384\), depth 10, dilation cycle \(1,2,4,8,16\) repeated twice, kernel size 15, dropout 0.2, one output channel; 3.13M parameters. \\
ETS-InceptionPyramid & Multi-scale temporal control & \(c_0=288\), branch width 96, depth 6, temporal scales 0.05/0.10/0.25/0.50~s, stem kernel 7, refinement kernel 11, dropout 0.2, one output channel; 3.25M parameters. \\
ETS-AttnSeg & Attention-convolution control & \(c_0=128\), depth 10, eight attention heads, convolution kernel 15, feed-forward and convolution expansion 2, dropout 0.2, attention dropout 0.1, one output channel; 3.05M parameters. \\
ETR-CNN large & Strongest scalar baseline & \(c_0=96\), widen 3, kernel size 15, dropout 0.2, high-resolution depth 12, low-resolution depth 10, three low-resolution stacks, two downsampling stages; 12.87M parameters. \\
ETR-CNN & Scalar temporal-readout baseline & Same family as ETR-CNN large with \(c_0=64\); 5.91M parameters. \\
MSP-CNN & Segment-pooling scalar baseline & \(c_0=48\), widen 2, depth per stage 3, kernel size 15, dropout 0.05; 3.01M parameters. \\
\bottomrule
\end{tabularx}
\end{table}

External EEG architectures are instantiated from the Braindecode constructors \citep{braindecode} named in the YAML configuration files and trained from scratch under the same data split, target support, optimizer, augmentation, checkpoint-selection, and holdout-evaluation protocol. Seed-level summaries, selected readout temperatures, prediction files, and reported aggregate tables are stored under \href{https://github.com/sneddy/neurosned/tree/main/benchmarks/experiments}{\texttt{benchmarks/experiments/}}.

Upon publication, we will release the benchmark materials supporting this study, including YAML configurations, data-preparation scripts, training and evaluation runners, model and loss code, selected readout temperatures, seed-level summaries, prediction files, shifted-crop diagnostics, aggregate tables, and figure-generation scripts. When raw HBN-EEG redistribution is restricted by source dataset terms, the release will provide the preparation code and expected split structure needed to reconstruct the CCD benchmark from the released data.

\FloatBarrier

\section*{Acknowledgments}

We would like to acknowledge support from NSF Award No.~2229881 for the AI Institute for Societal Decision Making (AI-SDM), the National Institutes of Health (NIH) under Contract R01HL159805, the MBZUAI-WIS Joint Program, the MBZUAI Startup Fund, and the Al Deira Causal Education project.

\end{document}